\documentclass[sn-mathphys-num]{sn-jnl}

\usepackage{graphicx}
\usepackage{subcaption}
\usepackage{multirow}%
\usepackage{amsmath,amssymb,amsfonts}%
\usepackage{amsthm}%
\usepackage{mathrsfs}%
\usepackage[title]{appendix}%
\usepackage{xcolor}%
\usepackage{textcomp}%
\usepackage{manyfoot}%
\usepackage{booktabs}%
\usepackage{algorithm}%
\usepackage{algorithmicx}%
\usepackage{algpseudocode}%
\usepackage{listings}%
\usepackage{natbib}%

\begin{document}

\title[Demystifying Functional Random Forests]{\textbf{Demystifying Functional Random Forests: Novel Explainability Tools for Model Transparency in High-Dimensional Spaces}}

\author*[1]{\fnm{Fabrizio} \sur{Maturo}}\email{fabrizio.maturo@unimercatorum.it}

\author[2]{\fnm{Annamaria} \sur{Porreca}}\email{annamaria.porreca@unimercatorum.it}
\equalcont{These authors contributed equally to this work.}

\affil*[1]{\orgdiv{Department of Economics, Statistics and Business}, \orgname{Faculty of Technological and Innovation Sciences}, \orgaddress{\street{Piazza Mattei, 10}, \city{Rome}, \postcode{66100}, \country{Italy}}}

\affil[2]{\orgdiv{Department of Economics, Statistics and Business}, \orgname{Faculty of Economics and Law}, \orgaddress{\street{Piazza Mattei, 10}, \city{Rome}, \postcode{66100}, \country{Italy}}}

\abstract{The advent of big data has raised significant challenges in analysing high-dimensional datasets across various domains such as medicine, ecology, and economics. Functional Data Analysis (FDA) has proven to be a robust framework for addressing these challenges, enabling the transformation of high-dimensional data into functional forms that capture intricate temporal and spatial patterns. However, despite advancements in functional classification methods and very high performance demonstrated by combining FDA and ensemble methods, a critical gap persists in the literature concerning the transparency and interpretability of black-box models, e.g. Functional Random Forests (FRF). In response to this need, this paper introduces a novel suite of explainability tools to illuminate the inner mechanisms of FRF. We propose using Functional Partial Dependence Plots (FPDPs), Functional Principal Component (FPC) Probability Heatmaps, various model-specific and model-agnostic FPCs' importance metrics, and the FPC Internal-External Importance and Explained Variance Bubble Plot. These tools collectively enhance the transparency of FRF models by providing a detailed analysis of how individual FPCs contribute to model predictions. By applying these methods to an ECG dataset, we demonstrate the effectiveness of these tools in revealing critical patterns and improving the explainability of FRF.}

\keywords{Functional Data Analysis, Random Forests, Explainability, High-Dimensional Data, Functional Principal Components, Model Transparency, Supervised Classification, Feature Importance}



\maketitle


\section{Introduction}

In the era of big data, analyzing high-dimensional datasets has become a critical challenge across various fields, including medicine, ecology, and economics. The explosion of data availability has provided outstanding opportunities for insight, but it has also introduced significant analytical challenges, mainly when dealing with the curse of dimensionality. This phenomenon, where the volume of data points grows exponentially with the number of dimensions, complicates extracting meaningful patterns and relationships using traditional statistical methods. Furthermore, when data are observed over time or across spatial dimensions, these challenges intensify due to the interconnected nature of the observations.

Functional Data Analysis (FDA) has emerged as a powerful solution to these challenges by transforming high-dimensional data into functional forms, which can capture complex temporal and spatial patterns more coherently and interpretably \citep{Ferraty2003, Ramsay1991}. FDA treats data as continuous curves rather than discrete points, enabling data analysis that varies over a continuum \citep{Ramsay1991, Shang2013}. By focusing on the underlying functional structure, FDA reduces dimensionality while preserving essential information, making it a robust tool for exploring complex datasets \citep{Febrero2012, Cuevas2014b, Ferraty2006}.

Over the past two decades, FDA has seen considerable advancements and has been applied successfully in a wide range of fields, particularly in biomedical applications. For instance, FDA has been used to analyze electrocardiogram (ECG) data, where the continuous monitoring of heart activity over time results in functional data well-suited to FDA techniques (see e.g.,  \citep{Maturo2022SIM}). In these applications, FDA helps not only reduce dimensionality but also uncover critical patterns that are crucial for accurate classification and diagnosis.

In the FDA literature, there are many supervised classification methods, mostly obtained as extensions of classification methods to the case of high dimensionality. In recent literature, also thanks to the success of machine learning methods in terms of performance, a new line of research is emerging that combines FDA and ensemble methods.
One of the most promising methods for the supervised classification of functional data is the Functional Random Forest (FRF), which integrates the flexibility of random forests with the analytical strengths of FDA \citep{Maturo2022SIM, maturo2024combining}. FRFs have demonstrated impressive performance in handling high-dimensional functional data for many applications. However, despite their effectiveness, FRF is a black-box model. Despite the excellent performances in accuracy, sensitivity, specificity, and AUC, today, from a statistical point of view, it leaves some doubts about the difficulties in explaining the results.
The need for transparency and interpretability in machine learning models is increasingly recognized, especially in critical fields such as healthcare and environmental science, where decisions based on model predictions can have far-reaching consequences. This is why medical doctors or entrepreneurs often prefer simpler but more interpretable methods.
For these reasons, the debate between black-box and transparent-box models remains a central and dynamic topic in contemporary statistical and computer science literature. In the era of big data, the quest for adequate interpretability and explainability methods has become one of the most pressing and significant challenges.

Focusing on the case of random forest for functional data, trained on FPCs (or equivalently on the coefficients of a fixed bases expansion),
the lack of transparency makes it difficult to interpret how individual features contribute to the model's predictions.
 In response to this need, this paper presents a novel set of explainability tools designed to enhance the interpretability of FRF. These tools aim to bridge the gap between model accuracy and understanding, providing users with the means to demystify the decision-making process of FRFs.

The key contributions of this work include the development of \textit{Functional Partial Dependence Plots (FPDPs)}, which illustrate the influence of individual FPCs on the model's predictions, and \textit{FPC Probability Heatmaps}, which visualize straightforwardly how changes in FPC scores affect the predicted probabilities for different classes. Besides, we present various model-specific and model-agnostic FPC importance metrics, including the \textit{FPC Internal-External Importance and Explained Variance Bubble Plot}, which provides a comprehensive view of the significance of each FPC from both internal and external perspectives.

These methods are demonstrated through an application to an ECG dataset, a classic example of functional data where the shape and variability of heart activity curves are crucial for accurate diagnosis. By applying these explainability tools, we show that it is possible to gain deep insights into how individual FPCs influence the classification of ECG signals, thereby enhancing the understanding of the FRF model. 
In summary, this paper makes a fascinating contribution to the FDA field by addressing a critical gap in the explainability of FRF. The novel explainability tools introduced here offer a pathway to greater transparency, making these powerful models more accessible and trustworthy for practitioners across various domains. 

The remainder of this paper is structured as follows. Section 2 outlines the foundational concepts of FDA, Functional Classification Trees (FCTs), and FRF. Section 3 presents the innovative explainability tools developed in this study, such as FPDPs and the FPC Internal-External Importance and Explained Variance Bubble Plot, among others. Section 4 applies these methods to an ECG dataset, demonstrating their practical utility and explaining their functionality in detail. Finally, Section 5 concludes with a discussion of the results and potential directions for future research.

\section{Material and methods preliminaries}

\subsection{Functional Data Analysis (FDA)}

Functional Data Analysis (FDA) focuses on treating data as functions, viewing each function as a single entity. In practice, however, these functional data are often observed as discrete points, reducing the function \( z = f(x) \) to a series of discrete observations, represented as pairs \( (x_j, z_j) \) where \( x_j \in \Re \) and \( z_j \) are the function values at the points \( x_j \), for \( j = 1, 2, \dots, T \) \citep{Ramsay2005}.

In FDA, we consider a functional variable \( X \) as a random variable that takes values in a functional space \( \Xi \), with a functional dataset being a sample \( \{x_1, \dots, x_N\} \), denoted as \( x_1(t), \dots, x_N(t) \), drawn from this functional variable \( X \) \citep{Ferraty2003}.
When considering a Hilbert space with a metric \( d(\cdot,\cdot) \) associated with a norm, where \( d(x_1(t), x_2(t)) = \|x_1(t) - x_2(t)\| \) and the norm \( \|\cdot\| \) is linked to an inner product \( \langle \cdot,\cdot \rangle \), a specific case is the space \( \mathcal{L}_2 \), which consists of square-integrable functions defined on \( \tau \). In this space, if \( x(t) \in \mathcal{L}_2 \), a basis system \( \phi_j(t) \) can be used to represent functions as a finite linear combination of basis functions.

The first step in FDA is converting observed values \( z_{i1}, z_{i2}, \dots, z_{iT} \) into a functional form. The most common method for estimating the functional data is using a fixed-basis approximation, where the function \( x_i(t) \) is approximated as:

\begin{equation}
x_i(t) \approx \sum_{s=1}^S c_{is} \phi_s(t),
\label{smoothfun}
\end{equation}

\noindent where \( c_{is} \) is the coefficient in the linear combination of the basis functions \( \phi_s(t) \) and subject $i$.

Alternatively, Functional Principal Components (FPCs) offer a data-driven approach that reduces dimensionality while effectively preserving the essential information content of the data. The functional data can then be approximated as:

\begin{equation}
x_i(t) \approx \sum_{k=1}^K \nu_{ik} \xi_k(t),
\label{fpca}
\end{equation}

\noindent where \( \nu_{ik} \) are the FPC scores, and \( \xi_k(t) \) are the principal components. By truncating this representation to the first \( p \) FPCs, the sample curves can be approximated, with the explained variance given by \( \sum_{k=1}^p \lambda_k \), where \( \lambda_k \) is the variance of the \( k \)-th FPC.

FPCs decompose the functional data into a series of orthogonal components that capture the maximum variance in the data. Mathematically, for a functional observation $x_i(t)$, the FPC decomposition can be also expressed as:

\begin{equation}
x_i(t) = \mu(t) + \sum_{k=1}^{K} \nu_{ik} \xi_k(t) + \epsilon_i(t),
\label{fpca22}
\end{equation}

\noindent where 
$\mu(t)$ is the mean function across all observations,
$\xi_k(t)$ is the $k$-th FPC,
$\nu_{ik}$ are the scores associated with each principal component for the $i$-th observation,
$\epsilon_i(t)$ is the residual noise term.
The enormous advantage of using this decomposition in supervised classification lies in its ability to filter out noise, achieve dimensionality reduction, and produce orthogonal features. By doing so, it addresses many challenges associated with multicollinearity in machine learning, leading to more robust and interpretable models.

Proximity measures between functions are crucial in FDA, with the \( \mathcal{L}_2 \)-distance being the most commonly used:

\begin{equation}
\left\| x_1(t)-x_2(t) \right\|_2 = \left\{ \frac{1}{\int_{\tau}w(t)\text{d}t} \int_{\tau} \left|x_1(t)-x_2(t)\right|^{2} w(t) \text{d}t \right\}^{1/2},
\label{dist}
\end{equation}

\noindent where \( w(t) \) is a positive weight function. Semi-metrics, such as the \( r \)-order derivative semi-metric, can also be used to capture more detailed information, and the FPC-based semi-metric is particularly useful for dimensionality reduction and interpreting similarity between functional data \citep{Ferraty2006, Ramsay2005, Febrero2012, Cuevas2014}.

\subsubsection{Functional Classification Trees (FCTs)}

In functional classification, the goal is to predict the class or label \( Y \) of an observation \( X \) by constructing a mapping \( f: \Xi \longrightarrow \{0, 1, \dots, U\} \), called a \textit{classifier}. This classifier assigns a predicted label to \( x \) with an error probability \( P \{ f(X) \neq Y \} \).

The continuous domain \( T \) can represent different aspects, such as time or space. While this study focuses on the time domain, the approach can be extended to other domains. The response could be categorical or numerical, leading to classification or regression problems. Here, we concentrate on scalar-on-function classification, particularly focusing on the so-called Functional Random Forest (FRF) \cite{Maturo2022SIM}.

Functional Classification Trees (FCTs) form the foundation of the FRF method, which is essentially an ensemble of weak FCTs. Therefore, to understand the FRF method fully, it is crucial to first illustrate how FCTs operate within the context of FDA.
Classification Trees (CT) are a supervised learning method used to predict a categorical response by learning decision rules from features \citep{Hyafil1976, Quinlan1986, Hastie2009}. They can be extended to the context of FDA by using the coefficients of a basis representation as features to train the functional classifier. This method is termed the Functional Classification Tree classifier and is fully illustrated in \citet{maturo2023supervised}, in particular regarding the meaning of the splitting rules in the functional context.

For a fixed basis system, such as that in Equation \ref{smoothfun}, the feature matrix \( \mathbf{C} \) is structured as follows:

\begin{equation}
\mathbf{C}=
\begin{pmatrix}
    c_{11} & \dots  & c_{1S}  \\
    \vdots & \ddots   &   \vdots \\
    c_{N1} &  \dots   & c_{NS}
\end{pmatrix},
\label{featuresspline}
\end{equation}

\noindent where \( c_{is} \) represents the coefficient of the \( i \)-th curve (\( i = 1,\dots,N \)) relative to the \( s \)-th basis function \( \phi_s(t) \).

Alternatively, for a data-driven basis system, as shown in Equation \ref{fpca}, the feature matrix \( \mathbf{V} \) is defined as:

\begin{equation}
\mathbf{V}=
\begin{pmatrix}
    \nu_{11} & \dots  &  \nu_{1K}  \\
    \vdots & \ddots   &   \vdots \\
     \nu_{N1} &  \dots   &  \nu_{NK} 
\end{pmatrix},
\label{featuresfpca}
\end{equation}

\noindent where \( \nu_{ik} \) is the score of the \( i \)-th curve (\( i = 1,\dots,N \)) corresponding to the \( k \)-th functional principal component \( \xi_{k} \) (\( k = 1,\dots,K \)).

In this study, we focus on FCTs trained using FPCs, which involves recursive binary partitioning of the feature space into regions (terminal nodes or leaves) composed of sets of functions \( x_i(t) \in X \).
To construct FCTs, the algorithm optimizes a cost criterion (e.g., Gini index or Shannon-Weiner entropy index) \citep{Hastie2009, rpart} at each step to achieve the best binary partition. The process begins with the complete set of FPCs' scores from Equation \ref{featuresfpca} and continues until terminal nodes are reached. Initially, a large FCT is generated, then pruned to balance complexity and accuracy.

As the FPCs' scores are used as features to predict \( Y \), interpreting FCTs differs from traditional CTs. The split values must be interpreted based on the domain portions represented by specific FPCs \( \xi_k(t) \) and the score thresholds. The scores' thresholds  determine whether curves belong to one subset or another based on whether the score \( \nu_{ik} \) is above or below this threshold (further details in \citep{Maturo2022SIM, maturo2024combining}).

\subsubsection{Functional Random Forest (FRF)}
\label{sec:223}

Classical Random Forest (RF) \citep{Ho1998} is a powerful machine learning algorithm, which is an extension of the bagging technique applied to decision trees. In RF, at each split during the tree-building process, a random subset of predictors is selected from the entire set. This random selection helps to decorrelate the trees and reduce the overall model variance, improving upon standard bagging where all trees could be dominated by the same strong predictor.
In FDA, a similar issue arises in bagging using FPCs because even though the functional dataset is bootstrapped, the resulting FCTs of the ensemble can be highly correlated. This is because the same FPCs often dominate the top splitting rules across multiple trees, leading to a lower reduction in variance.

FRF addresses this issue by further decorrelating the trees. Instead of using all available FPCs at each split, FRF randomly selects a subset of \( m \) FPCs as candidates for splitting, from the total set of \( K \) FPCs. When \( m < K \), the method is referred to as FRF; when \( m = K \), FRF reduces to bagging. By reducing the correlation between FCTs, FRF achieves greater variance reduction and more robust predictions.
In practice, a common heuristic is to set \( m \approx \sqrt{K} \). This ensures that, on average, a significant portion of FPCs are not considered at each split, further decorrelating the trees and improving the reliability of the averaged predictions from the forest.

In FRF, the prediction for an observation \( x_i(t) \) is made by aggregating the predictions from all the individual FCTs within the ensemble. The final decision is typically made by majority voting, where the most frequent class label predicted by the FCTs is selected as the final prediction.
The decision rule for predicting the class label \( \delta \) for an observation \( x_i(t) \) can be expressed as:

\begin{equation}
\hat{f}^{(m)}_{rf}(x_i(t)) = \text{mode} \left\{ \hat{f}^{(1)}(x_i(t)), \hat{f}^{(2)}(x_i(t)), \dots, \hat{f}^{(M)}(x_i(t)) \right\}
\end{equation}

\noindent where 
     \( \hat{f}^{(m)}_{rf}(x_i(t)) \) is the predicted label by the FRF for the observation \( x_i(t) \),
     \( \hat{f}^{(j)}(x_i(t)) \) is the predicted label by the \( j \)-th FCT in the ensemble,
     \( M \) is the total number of trees in the ensemble,
     and the mode function selects the most frequently predicted class among the predictions from all \( M \) trees.

Unlike individual FCTs, which might require pruning to prevent overfitting, FRF benefit from the combined effects of bootstrap aggregation and feature selection at each split, reducing the likelihood of overfitting even without pruning. Consequently, in FRF, FCTs are not pruned, in contrast to the approach used for individual FCTs discussed earlier.

The key to the success of FRF lies in the diversity of the individual FCTs in the ensemble. This diversity is achieved through double randomization of units and features. By reducing the correlation between the FCTs, FRF ensures that each FCT captures different aspects of the data, leading to a more robust and accurate final model, improving its generalization ability.

\section{Novel Explainability Tools for Functional Random Forest}

\subsection{The Intuition Behind the Proposed Explainability Tools}

The intuition behind explainability using Functional Partial Dependence Plots (FPDPs) and Functional Principal Component Probability Heatmap (FPCPH), in the FDA context, lies in the ability to study the behaviour of reconstructed curves through a single FPC, while keeping all other components fixed. This approach allows us to observe how variations in the coefficient of the linear combination affect the functional data reconstruction.

In other words, we can analyze the behaviour of curves' reconstruction when only one component’s score is altered while the scores of the other components remain constant. 
Similarly, another possibility is to study what happens to the curve when it is reconstructed using only one FPC at a time, varying the score of that single component while setting all other scores to zero. This approach allows us to isolate and understand the specific contribution of each FPC to the overall shape of the functional data. By analyzing the curve in this manner, we can gain insights into how individual FPCs influence the reconstructed data and how their variations impact the overall functional form.

When reconstructing the functional data \( x_i(t) \) using only a single FPC, say the \( k \)-th component, and ignoring all other components, the approximation \( \hat{x}_i(t; \nu_{ik}) \) can be expressed as:

\begin{equation}
\hat{x}_i(t; \nu_{ik}) = \nu_{ik} \xi_k(t),
\label{sole}
\end{equation}

\noindent where 
     \( \nu_{ik} \) is the score associated with the \( k \)-th FPC, and
     \( \xi_k(t) \) is the \( k \)-th FPC.
It follows that the reconstruction \( \hat{x}_i(t; \nu_{ik}) \) in Equation \ref{sole} considers solely the contribution of the \( k \)-th FPC, without involving the other  FPCs.
This simplified representation is valuable for understanding how the variation in a single score \( \nu_{ik} \) affects the shape of the reconstructed function \( \hat{x}_i(t) \). By varying \( \nu_{ik} \) while assuming the other scores are zero (i.e., \( \nu_{ij} = 0 \) for all \( j \neq k \)), we can observe the isolated effect of the \( k \)-th FPC \( \xi_k(t) \) on the reconstruction of the curve \( x_i(t) \).

Mathematically, this means that the reconstruction of the functional data \( x_i(t) \) is entirely determined by the \( k \)-th principal component \( \xi_k(t) \) and its corresponding score \( \nu_{ik} \). As \( \nu_{ik} \) varies, the shape of the function \( \hat{x}_i(t) \) will vary according to the \( k \)-th FPC \( \xi_k(t) \).
This concept is crucial when moving on to interpret the explainability measures in the subsequent subsections, where the goal is to understand how individual FPCs contribute to the variability of the response variable. The isolated variation of a single FPC provides a clear understanding of how each component can influence the model output, thereby forming the basis for interpreting explainability metrics and tools.

\begin{figure}[htbp]
    \centering
    \begin{subfigure}[b]{0.45\textwidth}
        \centering
        \includegraphics[width=\textwidth]{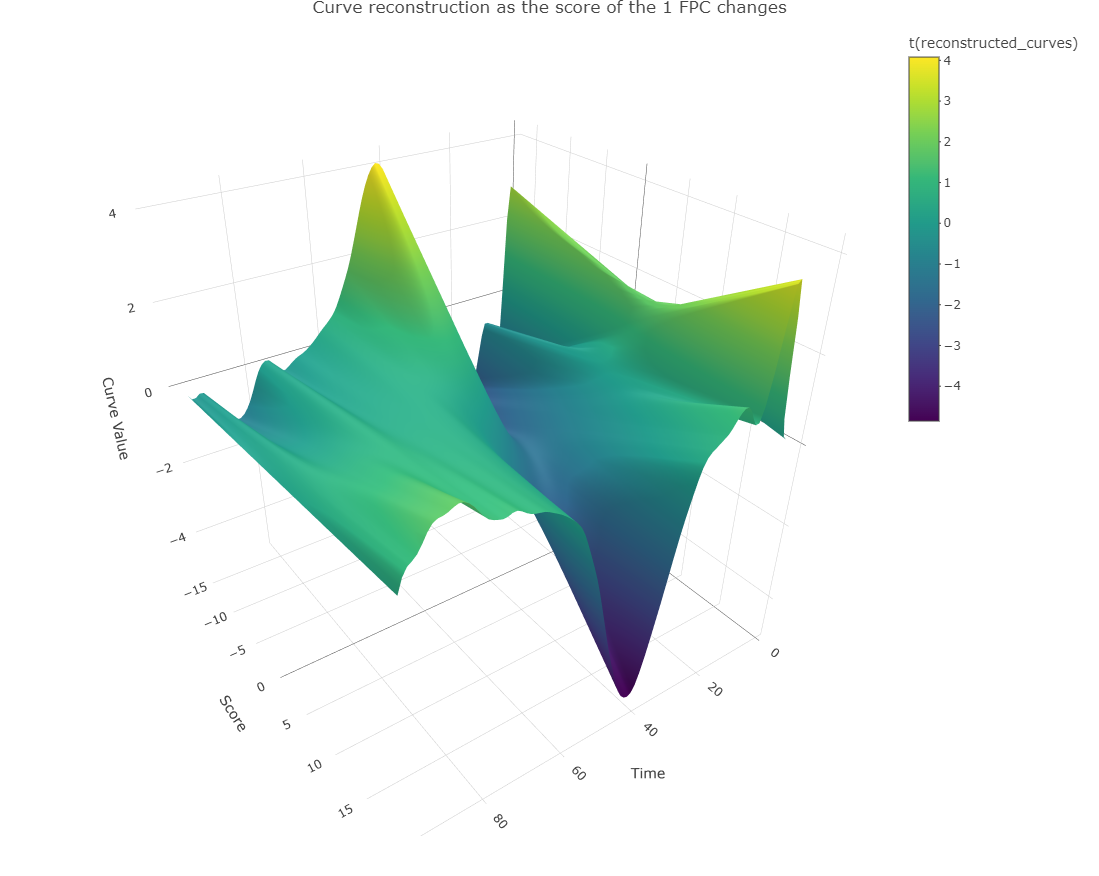}
        \caption{First FPC variation.}
        \label{fig:fpc_variation1}
    \end{subfigure}
    \hfill
    \begin{subfigure}[b]{0.45\textwidth}
        \centering
        \includegraphics[width=\textwidth]{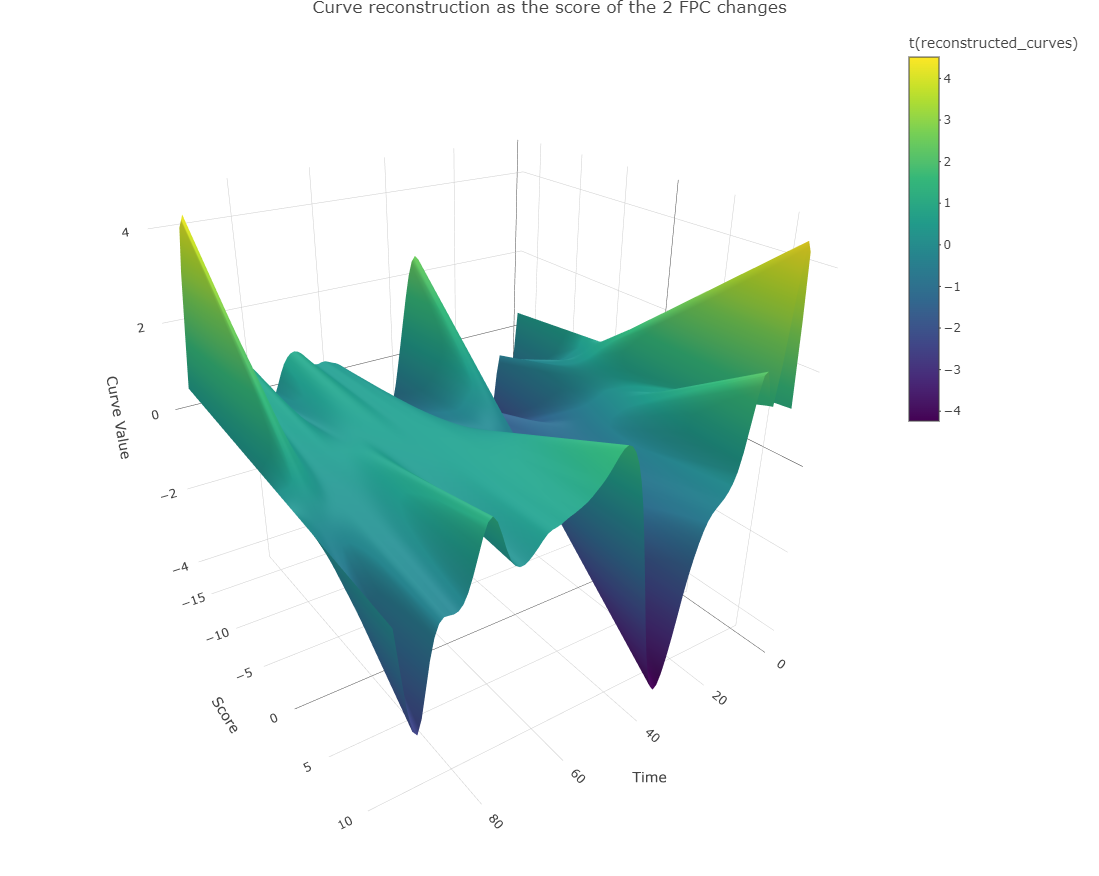}
        \caption{Second FPC variation.}
        \label{fig:fpc_variation2}
    \end{subfigure}
    \vspace{0.5cm}
    \begin{subfigure}[b]{0.45\textwidth}
        \centering
        \includegraphics[width=\textwidth]{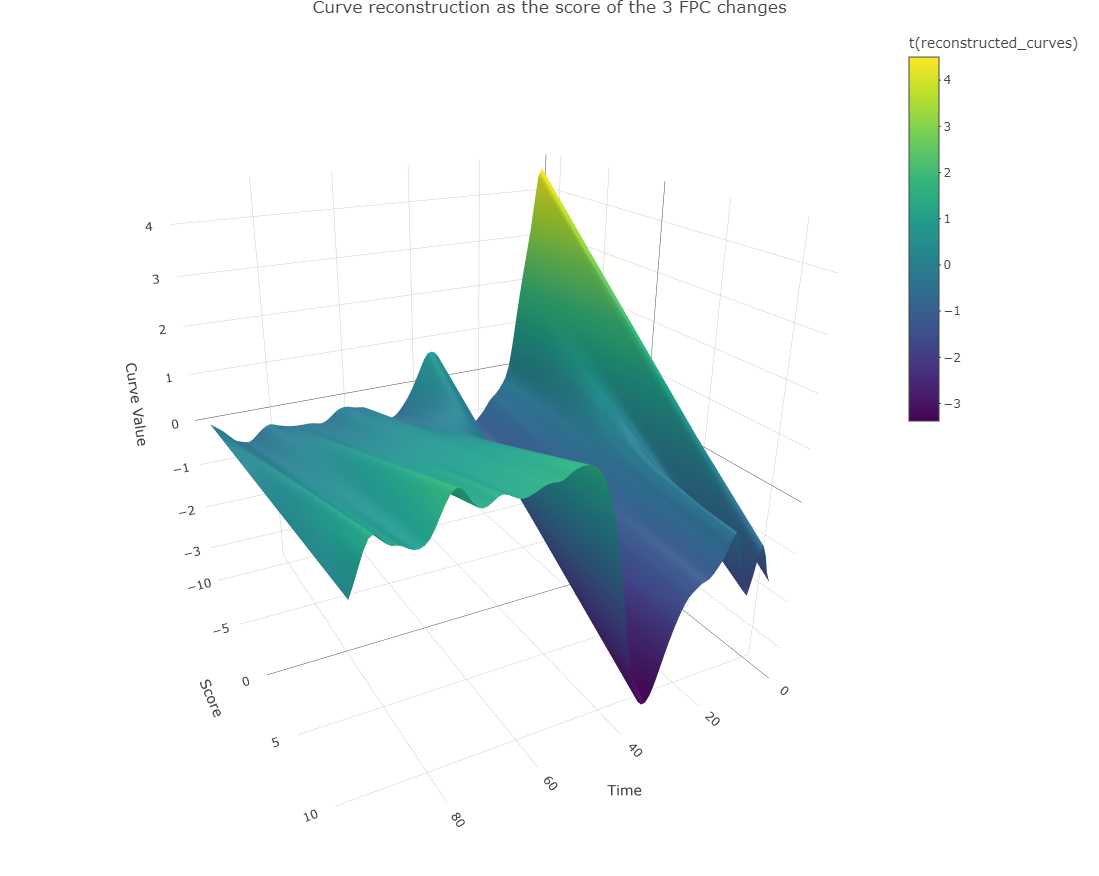}
        \caption{Third FPC variation.}
        \label{fig:fpc_variation3}
    \end{subfigure}
    \hfill
    \begin{subfigure}[b]{0.45\textwidth}
        \centering
        \includegraphics[width=\textwidth]{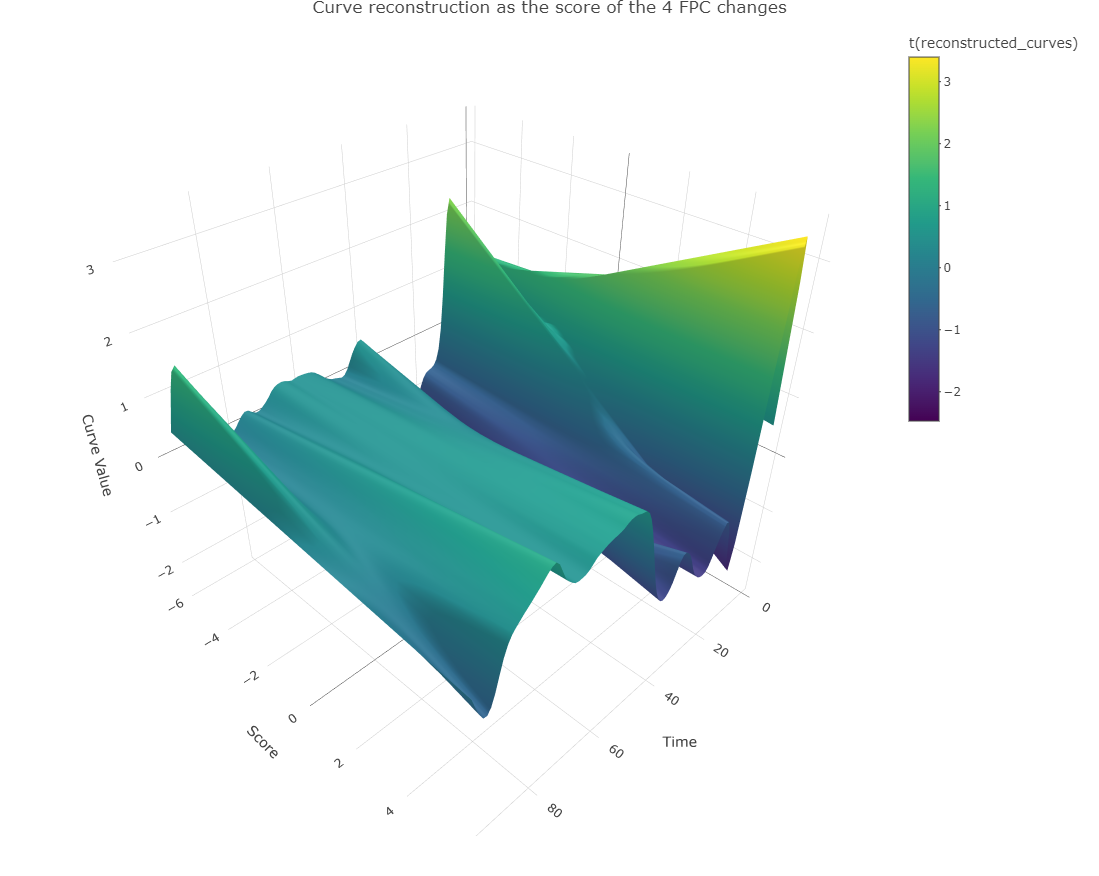}
        \caption{Fourth FPC variation.}
        \label{fig:fpc_variation4}
    \end{subfigure}
    \caption{Variation of the first four Functional Principal Components (FPCs) with changes in their scores.}
    \label{fig:fpc_variation}
\end{figure}

To illustrate this concept, we use the \textit{ECG200} dataset \citep{ecg200dataset}, which we will also employ in the application section. We compute FPCs on this dataset and show how the FPCs derived from the ECG data contribute to the reconstruction of the signals and how the variation in individual FPC scores affects the reconstructed curves.
Figure \ref{fig:fpc_variation} demonstrates this concept by displaying how, for example, the first four FPCs change as their scores vary, thereby providing insight into the contribution of each FPC to the overall variation in the data.
This visualization helps in understanding the impact of each FPC on the functional data, facilitating a clearer interpretation of the model’s behavior in terms of the principal components.

By isolating the effect of one component at a time, we can observe how it uniquely contributes to the overall shape and behavior of the functional data.
This understanding is crucial because it provides a foundation for interpreting more complex explainability measures. In the next section, these measures aim to go beyond the mere reconstruction of data; they seek to explain the contribution of specific features to the response variable in predictive models. Essentially, if we grasp how each FPC score affects the data reconstruction, we can extend this insight to understand how each feature influences the model's output.
This knowledge is not just theoretical but practical, particularly in fields where model transparency and interpretability are essential. For example, in predictive modeling, it's important to identify which features are driving predictions and to what extent. The ability to isolate and understand the impact of individual components or features enables us to build models that are not only accurate but also interpretable and trustworthy. 

\begin{figure}[htbp]
    \centering
    \includegraphics[width=12cm]{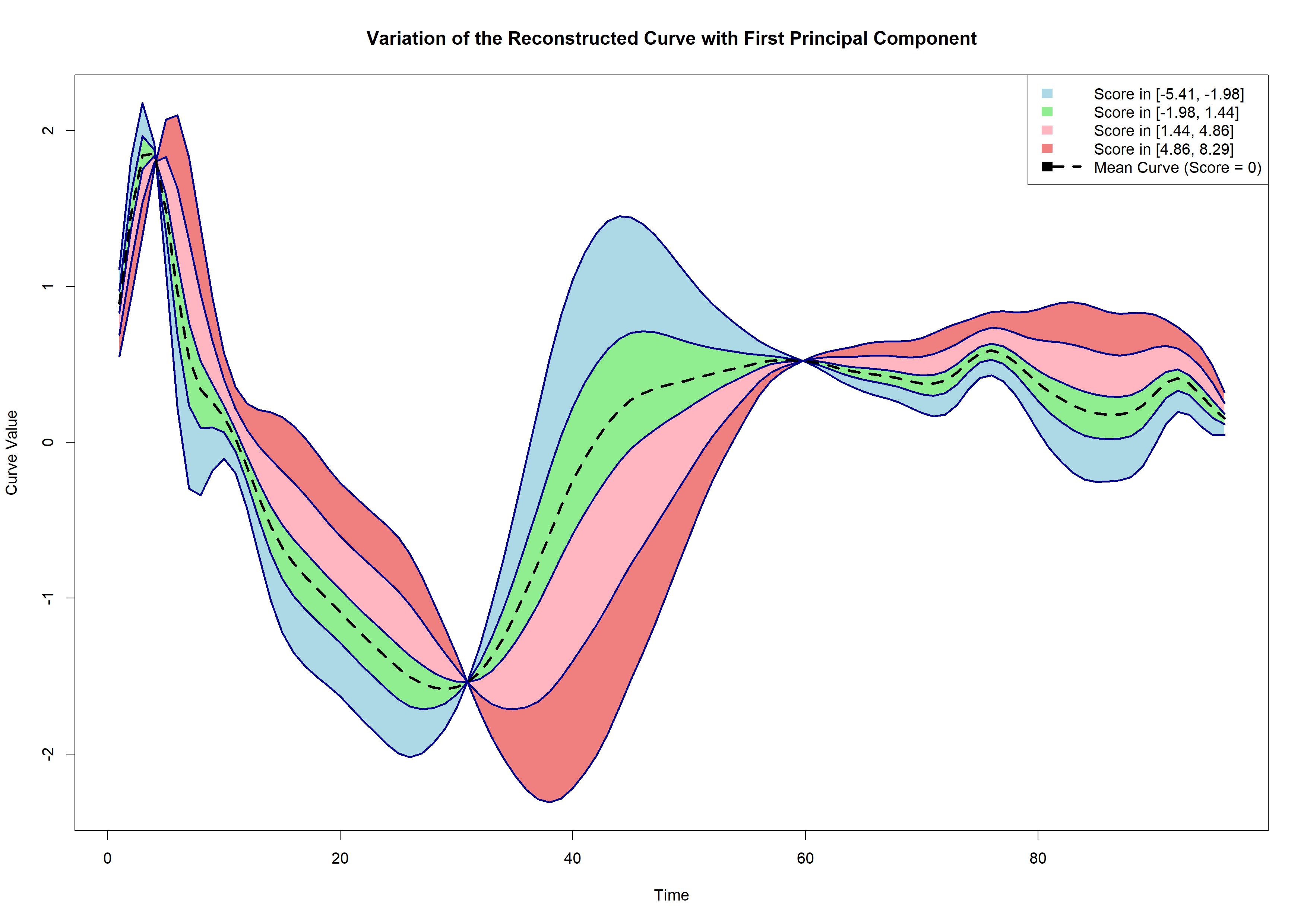}
    \caption{Variation of the reconstructed curves using the First Functional Principal Components (FPC) based on a categorization of the ranges of its score.}
    \label{fig:fpc_variation_2x2}
\end{figure}

Alternatively, to simplify the interpretation of how variations in a single FPC score affect the reconstructed curve, we can use 2D plots that stratify the curves based on ranges of FPC scores. Figure \ref{fig:fpc_variation_2x2} shows an example of this approach, where we visualize the first FPC.
The reconstructed curves are visualised to reflect the different intervals of the first FPC scores. The shaded regions highlight the range of variability within specific score intervals, providing a visual stratification of the FPC's influence on the functional data. This approach allows for a clearer interpretation of how variations in the FPC score can alter the reconstructed curve shapes across different regions of the domain.
The colored bands in the figure represent the potential range of curve values corresponding to each score interval. By focusing on these shaded areas, it becomes evident how the first FPC contributes to the overall shape of the functional data. For instance, the red band, representing higher score intervals, shows a significant deviation from the mean curve, indicating that higher scores have a stronger impact on the curve's amplitude and shape in the the central part of the domain, but not only. 
Moreover, the black dashed line represents the mean curve, where the FPCS scores are zero, serving as a reference point. The variation around this mean curve across different score intervals reveals how specific functional features are emphasized or diminished depending on the score, aiding in the interpretation of the FPC's role in the data.

\subsection{Functional Partial Dependence Plots (FPDPs)}

This research proposes the concept of Functional Partial Dependence Plots (FPDPs) and thus extends the concept of Partial Dependence Plots (PDPs) \citep{friedman2001greedy} to functional data. The aim is to visualise the marginal effect of a single FPC on the predicted probability of a binary outcome. 
FPDPs illustrate how changes in a specific FPC (while holding other features constant) influence the probability of the predicted class. This allows us to assess which ranges of the FPCs' score values increase or decrease the probability of belonging to a specific class.
These plots can help us understand the functional model's behavior and how individual FPCs contribute to the final prediction.

Classical PDPs are used to visualize the marginal effect of a single feature on the predicted outcome of a machine learning model, such as a random forest or gradient boosting machine. The idea behind PDPs is to show how the model’s predictions change as the value of a particular feature varies, while averaging out the effects of all other features. By fixing the feature of interest at different values and averaging the predictions across all possible values of the other features, PDPs reveal the relationship between that feature and the outcome, providing insight into the model's behavior.
Consider a model where the predicted outcome $\hat{y} = f(x_1, x_2, \dots, x_p)$ depends on a set of scalar features $x_1, x_2, \dots, x_p$. The PDP for a feature $x_j$ is calculated by averaging the model’s predictions over the joint distribution of all other features $x_{\setminus j}$. Mathematically, the partial dependence function for a feature $x_j$ is defined as:

\begin{equation}
\text{PDP}(x_j) = \frac{1}{N} \sum_{i=1}^{N} f(x_j, x_{\setminus j}^{(i)})
\end{equation}

\noindent where $N$ is the number of observations, and $x_{\setminus j}^{(i)}$ represents the values of all features except $x_j$ for the $i$-th observation. This function is plotted against the values of $x_j$ to visualize how changes in $x_j$ affect the average predicted outcome $\text{PDP}(x_j)$.

When extending the concept of Partial Dependence Plots (PDPs) to FDA, where the coefficients of FPCs represent the features, the mathematical approach remains largely similar, but the interpretation becomes distinct and more nuanced.
The process involves calculating and visualizing how the predicted outcome changes as we vary the score $\nu_{ik}$ of a specific FPC while averaging out the effects of the other FPC coefficients. This is done by selecting a range of values for $\nu_{ik}$ and fixing it at these values one at a time. For each fixed value, the model’s predictions are calculated by averaging across all other FPC coefficients. The results are then plotted, showing how the predicted probability changes with the varying coefficient.

This extension of FPDPs to FPC coefficients allows us to understand the marginal effect of individual components of the functional data on the model’s predictions. For instance, if the FPDP for a particular FPC's coefficient shows a steep increase, this would indicate that as this coefficient increases, the FPC it represents contributes significantly to increasing the predicted probability of a certain class. Conversely, a flat FPDP suggests that the coefficient has little to no effect on the prediction. This method is precious in FDA, where understanding the contribution of specific components to the model's output can be complex.
The goal of FPDPs is to show how the model's prediction $\hat{y}$ changes as we vary a specific FPC score $\nu_{ik}$ while averaging out the effects of all other FPC scores $\nu_{\setminus k}$. 

The partial dependence function for the score $\nu_{k}$ is given by:

\begin{equation}
\text{FPDP}(\nu_{k}) = \frac{1}{N} \sum_{i=1}^{N} f\left(\nu_{k}, \nu_{\setminus k}^{(i)}\right)
\end{equation}

\noindent where $\nu_{\setminus k}^{(i)}$ denotes all other FPCs' scores for the $i$-th observation except $\nu_k$, and the function $f(\nu_k, \nu_{\setminus k}^{(i)})$ gives the predicted outcome based on the scores. The FPDP is computed by varying the score $\nu_k$ over a range of values while holding all other scores $\nu_{\setminus k}$ at their observed values. The plot of $\text{FPDP}(\nu_k)$ versus $\nu_k$ reveals how changes in this particular FPC score influence the predicted outcome. 

When creating FPDPs, we have the option to use either the probability scale or the logit scale. The probability scale is intuitive and directly interpretable, representing the predicted probability of an event occurring. This makes it easy to understand and communicate the results. On the other hand, the logit scale, which expresses probabilities in terms of log-odds, can provide greater sensitivity to changes in the middle range of probabilities. This is particularly useful when analysing models where the distinction between different classes is subtle. The choice between the two depends on whether clarity or sensitivity is more critical for the analysis.

This extension of PDPs to FPCs' scores allows for interpreting the contribution of specific scores to the model's predictions.
Nevertheless, it is crucial to reconstruct a function across the range of score values using the corresponding FPC to go beyond interpreting the score's effect and gain a deeper understanding within the time domain. This approach allows us to discern the specific shapes and characteristics of the curves that lead to the prediction effects suggested by the FPDP. For this reason, the interpretation of the FPDP should always be accompanied by a graph of the function reconstructed with that specific FPC, as illustrated in Figure \ref{fig:fpc_variation_2x2}. In Section 4, the joint reading of these pictures will be described in detail.

\subsection{Functional Principal Components Probability Heatmap (FPCPH)}

The idea behind the Functional Principal Components Probability Heatmap (FPCPH) is to understand in a single graph the effect of the FPCs' scores values on the response variable. 
The proposal is to generate a range of values for each FPC's score while keeping the scores of other FPCs constant. Then, based on these scores, the FRF classifier predicts the probability of belonging to a specific class.
For each FPC \( \xi_k(t) \), a sequence of scores \( \nu_{ik} \) is defined over a range, and the predicted probability of the observation being classified into a particular class (e.g., Class 1) is calculated using the trained FRF. 

Statistically, a FRF is trained using the scores \( \nu_{ik} \) from FPCs as input features. The model aims to predict the class label \( Y_i \) for each observation \( i \), where \( Y_i \in \{0, 1\} \) in the case of binary classification. The classification model is built as follows:

\begin{equation}
\hat{Y}_i = f(\nu_{i1}, \nu_{i2}, \ldots, \nu_{iK}),
\end{equation}

\noindent where \( f(\cdot) \) is the function learned by FRF, mapping the FPCs' scores to the predicted class label \( \hat{Y}_i \).

To assess the impact of each FPC on the classification probabilities, we vary the score \( \nu_{ik} \) of each FPC while holding the other scores constant (e.g., at their mean values). For a given FPC \( k \), we generate a sequence of scores \( \nu_{ik} \) over a specified range \( [\nu_{\text{min}}, \nu_{\text{max}}] \):

\begin{equation}
\nu_{ik}^{(1)}, \nu_{ik}^{(2)}, \ldots, \nu_{ik}^{(M)},
\end{equation}

\noindent where \( M \) represents the number of score values sampled.
For each score \( \nu_{ik}^{(m)} \), the FRF is used to predict the probability of belonging to a specific class:

\begin{equation}
\hat{P}_{i}^{(m)} = \Pr(Y_i = 1 \mid \nu_{ik} = \nu_{ik}^{(m)}, \nu_{ij} = \bar{\nu}_{ij}, \forall j \neq k),
\end{equation}

\noindent where \( \bar{\nu}_{ij} \) represents the mean score for FPC \( j \).

The predicted probabilities \( \hat{P}_{i}^{(m)} \) are then used to construct a heatmap, where the x-axis represents different FPCs and the y-axis represents the range of scores for each FPC. The intensity of the color in the heatmap indicates the predicted probability of the class of interest:

\begin{equation}
\text{Heatmap}(k, m) = \hat{P}_{i}^{(m)},
\end{equation}

\noindent where \( k \) indexes the FPC and \( m \) indexes the score within the range for that FPC.

The results are then visualised as a heatmap (namely \textit{FPCPH}), where the x-axis represents different FPCs, the y-axis represents the range of scores for each FPC, and the colour intensity indicates the predicted probability of class one membership. This heatmap provides a transparent and interpretable visual representation of how changes in each FPCs' score affect the FRF predictions. 
By analysing the heatmap, one can determine which aspects of the functional data, as captured by the FPCs, are most critical for distinguishing between classes in the classification model.
To fully understand how score variations impact model predictions, especially in time-dependent data, it is essential to reconstruct the function across the entire range of score values for the relevant FPC. This allows us to identify the specific shapes and patterns that drive the effects shown by the FPDP. Therefore, similarly to FPDPs, the interpretation of the FPCPH should always be paired with a graph of the function reconstructed using that FPC, as shown in Figure \ref{fig:fpc_variation_2x2}.

\subsection{Ranking and Assessing Functional Principal Components Importance}
\label{modelagnospec}

In statistical modeling and machine learning, ranking variables based on their importance in predicting the outcome is crucial. This section describes different methods for ranking FPCs' importance, distinguishing between internal and external tools.

\subsubsection{Functional Principal Components' Internal Importance Measures (Model-Specific Feature Importance)}

\noindent  \textbf{Mean Decrease in Gini Impurity}\\

The simplest classic measure of explainability to extend to FRF, as presented in \citep{Maturo2022SIM}, is the Mean Decrease in Gini Impurity.
In general, Random Forests can inherently measure feature importance through metrics such as the Mean Decrease in Gini Impurity because it is part of the process of creating the individual constituents, i.e classification trees.

The Gini Index \( G(t) \) at a node \( o \) of a FCT can be calculated as:

\begin{equation}
G(o) = 1 - \sum_{i=1}^{C} p_i^2
\end{equation}

\noindent where \( C \) is the number of classes and \( p_i \) is the proportion of curves belonging to class \( i \) at node \( o \). The Mean Decrease in Gini (MDG) for a FPC \( \xi_k \) is computed by averaging the total decrease in Gini Index due to splits on \( \nu_k \) across all the trees in the forest:

\begin{equation}
\text{MDG}(\nu_k) = \frac{1}{T} \sum_{o=1}^{O} \sum_{s \in S_j^t} \Delta G_s^o
\end{equation}

\noindent where 
  \( O \) is the total number of FCTs in the forest,
   \( S_j^o \) is the set of all nodes in tree \( o \) where feature \( \nu_k \) is used for splitting,
  \( \Delta G_s^o \) is the decrease in Gini Index at node \( s \) in tree \( o \) due to the split on \(\nu_k \).\\[\baselineskip]

\noindent  \textbf{Permutation Importance}\\

Permutation Importance assesses the importance of a FPC by measuring the increase in the model’s prediction error when the values of that variable are randomly shuffled, thereby breaking the relationship between the variable and the outcome. For a given FPC \( \xi_k \), its importance is computed as:

\begin{equation}
\text{PI}(\nu_k) = \text{Error}_{\text{perm}}(\xi_k) - \text{Error}_{\text{orig}}
\end{equation}

\noindent  where \( \text{Error}_{\text{perm}}(\nu_k) \) is the prediction error after shuffling \( \nu_k\), and \( \text{Error}_{\text{orig}} \) is the original prediction error. A higher difference indicates greater importance, as the model’s performance deteriorates more when an important FPC is perturbed.

\subsubsection{Functional Principal Components' External Importance Measures and Tools (Model-Agnostic Feature Importance)}

\textbf{FPCs Scores Distribution Conditioned on the Response}\\

The basic idea of an external measure is to understand how much a FPC helps explain a difference between outcome categories.
Model-agnostic importance measures are the typical explainability measure because they are external tools that try to help us understand what a back-box model hides and do not depend on the model.

Hence, the primary objective is to identify which FPCs can contribute to the predictive model by analysing their variability to the response.
The simplest thing to do as a preliminary analysis is certainly to examine the conditional distribution of FPC scores based on the response variable.
Let \(\nu_{ik}\) denote the score of the \(i\)-th observation for the \(k\)-th FPC. Consider the binary response variable \(Y\) with two classes \(Y = 0\) and \(Y = 1\). The conditional distribution of \(\nu_{ik}\) given \(Y = y\) can be explored using violin plots, which allow us to compare the distributions of FPC scores across the two classes and combine the benefits of a boxplot and a density plot:

\begin{equation}
f(\nu_{ik} \mid Y = y), \quad y \in \{0, 1\}
\end{equation}

\noindent \textbf{F statistic}\\

To quantitatively assess the relationship between the FPCS' scores and the binary response variable, we can use an Analysis of Variance (ANOVA). ANOVA allows us to test whether the mean FPC scores differ significantly between the two response classes. A significant ANOVA result indicates that the FPC's score is related to the response variable and might be an influential feature in the predictive model.

Given the FPC scores \(\nu_{ik}\) and the binary response \(Y_i\), the ANOVA model can be extended to this context as follows:

\begin{equation}
\nu_{ik} = \mu_k + \alpha_k Y_i + \epsilon_{ik}
\end{equation}

\noindent where \(\mu_k\) is the overall mean of the \(k\)-th FPC score, \(  \alpha_k\) represents the effect of the binary response variable on the \(k\)-th FPC score, and \(\epsilon_{ik}\) is the error term. The hypothesis tested in ANOVA is:

\begin{align*}
H_0: & \quad \alpha_k = 0 \quad \text{(no difference in FPC scores between classes)} \\
H_1: & \quad \alpha_k \neq 0 \quad \text{(significant difference in FPC scores between classes)}
\end{align*}

As in the classical case, for ANOVA to yield robust results, the fundamental assumptions must be met: independence of observations, normality of residuals, and homogeneity of variances across groups. If these assumptions are violated, the validity of the results may be affected.
A significant ANOVA result indicates that the FPC is associated with the response variable, suggesting its potential importance in the predictive model.
Given an FPC coefficient \( \nu_{ik} \) associated with the \( k \)-th FPC, the ANOVA F-statistic is used to quantify the contribution of \( \nu_{ik} \) to the variability in the response variable \( Y \). The F-statistic for \( \nu_{ik} \) is defined as:

\begin{equation}
F_k = \frac{\text{MS}_{\text{model}, k}}{\text{MS}_{\text{error}}}
\end{equation}

\noindent where
\(\text{MS}_{\text{model}, k}\) represents the mean square for the model associated with the \( k \)-th FPC coefficient \( \nu_{ik} \). It is calculated as:

    \[
    \text{MS}_{\text{model}, k} = \frac{\text{SS}_{\text{model}, k}}{\text{df}_{\text{model}, k}}
    \]

 \noindent    where \(\text{SS}_{\text{model}, k}\) is the sum of squares due to the model, and \(\text{df}_{\text{model}, k}\) is the degrees of freedom for the model associated with \( \nu_{ik} \).

\(\text{MS}_{\text{error}}\) is the mean square of the residual error, calculated as:

    \[
    \text{MS}_{\text{error}} = \frac{\text{SS}_{\text{error}}}{\text{df}_{\text{error}}}
    \]

\noindent     where \(\text{SS}_{\text{error}}\) is the sum of squares of the residuals, and \(\text{df}_{\text{error}}\) is the degrees of freedom associated with the error.

The F-statistic \( F_k \) tests the null hypothesis that the FPC's score \( \nu_{ik} \) (response variable) is not affected by the grouping variable \( Y \) (response variable in FRF but not in the ANOVA context). A larger \( F_k \) value suggests a higher importance of that FPC.\\[\baselineskip]

\noindent  \textbf{\(\eta^2\) statistic}\\

To quantify the effect size of each FPC, we can use the \(\eta^2\) statistic, which measures the proportion of the total variance in the response explained by the FPC coefficient \( \nu_{ik} \). The \(\eta^2\) for \( \nu_{ik} \) is defined as:

\begin{equation}
\eta^2_k = \frac{\text{SS}_{\text{model}, k}}{\text{SS}_{\text{total}}}
\end{equation}

\noindent where
 \(\text{SS}_{\text{model}, k}\) is the sum of squares attributed to the model associated with the \( k \)-th FPC coefficient \( \nu_{ik} \), reflecting the variance explained by this coefficient.
 \(\text{SS}_{\text{total}}\) is the total sum of squares, which represents the total variance in the response variable \( Y \).
 
The \(\eta^2_k\) value can provide a measure of the importance of each FPC coefficient \( \nu_{ik} \), indicating how much of the total variance in \( Y \) is explained by \( \nu_{ik} \). Higher \(\eta^2_k\) values suggest that the corresponding FPC should have a greater impact on the response, despite what happens in FRF. 

While the F-statistic is primarily used for hypothesis testing, \(\eta^2\) offers a descriptive measure of effect size, providing valuable insight into the magnitude of each FPC's contribution to the prediction.

\subsection{Model-Specific VS Model-Agnostic Feature Importance: the FPCs Internal-External Importance and Explained Variance Bubble Plot}

Understanding the importance of input features is critical for model explainability in predictive modelling, particularly with complex models like FRF. 
Each method proposed in Section \ref{modelagnospec} offers a different perspective on FPCs importance. 
Internal measures capture how much each FPC contributes to FRF's predictive accuracy, but they are model-dependent; internal measures reflect the importance of FPCs in the context of how FRF interprets and utilises them. The MDG from FRF offers an ensemble method’s view, emphasising how well each variable splits the data. Indeed, FRF inherently provides MDG because it is connected to the impurity in each FCT node and aggregates the impurity reduction across all trees in the FRF.
Permutation Importance, on the other hand, stresses how sensitive the model's performance is to each FPC, delivering a direct measure of the FPC's contribution to predictive accuracy.

While model-specific metrics are helpful, they provide a view restricted to the algorithm employed. To gain a more comprehensive understanding, it is beneficial to compare these internal measures with external measures derived from simpler, interpretable models, like ANOVA. 
ANOVA-based metrics can evaluate the importance of FPCs via statistical information on the variance explained by each FPC. Thus this approach is useful in understanding the effect size or the ratio of the variability between groups to the variability within groups, producing metrics such as the F-statistic and Eta Squared (\(\eta^2\)). These metrics offer a model-agnostic perspective on FPCs importance, reflecting the inherent predictive power of a FPC independently of the complexities of the predictive model.

By visualising the importance of FPCs using internal and external metrics, we can identify FPCs that are universally important across different modeling perspectives and those whose importance is more model-specific. This dual approach helps in several ways.
Critical FCPs identification involves recognising FPCs that score highly on internal and external metrics. These FPCs are likely essential to the predictive model as they drive its performance and possess strong, independent predictive power, making them reliable outcome indicators.
Model-specific bias occurs when FPCs are crucial internally but not externally, indicating possible model-specific biases. The model's RFR mechanism might overestimate these FPCs despite the fact that they are not genuinely predictive features; the latter check is vital for detecting overfitting or spurious correlations.
Understanding external validity is highlighted when features are important externally but not internally, suggesting that the model might be underutilising important information. This discrepancy could signal areas for model improvement, where adjustments to the model's structure might be necessary to grasp the full predictive potential of these features.

The presence of FPCs explaining much variability does not guarantee that they are essential for supervised classification tasks. Indeed, the features explaining less variability than the first FPCs are often decisive for achieving high performance since, particularly the first FPCs, usually capture a variability common to all curves.
Thus, to integrate the information just described with the information on variability explained within the decomposition into FPCs, we propose the last explainability tool, namely the \textit{FPCs internal-external importance and explained variance bubble plot}.
This instrument exploits quadrant analysis and allows us to categorise FPCs by plotting importance metrics in a two-dimensional space, with external importance on one axis and internal importance on the other (clearly, we can select one internal and one external based on the previously proposed metrics). 
This approach enables the classification of FPCs into quadrants, providing a nuanced interpretation of each FPC's role in both the model and the data.
This approach underscores the importance of cross-verifying the model-specific interpretations with more generalised, model-agnostic assessments of feature importance. It helps ensure that the model is not just fitting the data but also relying on genuinely predictive features, thus enhancing its explainability and robustness.

\section{Application}

To illustrate the proposed instruments, we utilise the ECG200 dataset, initially proposed by  Olszewski at Carnegie Mellon University in 2001 as part of his work titled “\textit{Generalized feature extraction for structural pattern recognition in time-series data}” \citep{ecg200dataset}. The ECG200 dataset is a benchmark for testing new classifiers and consists of time-series data, where each series represents the electrical activity recorded during a single heartbeat. The dataset is divided into two classes: Normal Heartbeat (NH) and Myocardial Infarction (MI). It contains 100 signals in the training set and 100 signals in the test set, making it a compact yet challenging dataset for classification tasks.
The data is publicly accessible at https://www.timeseriesclassification.com/description.php?Dataset=ECG200. The primary objective is to predict whether a new patient, based on their observed ECG, is healthy or suffering from a myocardial infarction. 

This prediction task provides a practical example to demonstrate the effectiveness of our proposed method in handling functional data.
We stress that in this study we are not interested in the performance of the model, which other studies have already extensively verified and compared with other classifiers (see \citep{Maturo2022SIM, maturo2023supervised}). In this context, the aim is to show how to use the proposed explainability tools for demystifying the FRF.
For this reason, comparison with other functional classifiers, or evaluations of performance measures are not the purpose of the paper. Similarly, the comparison with other explainability tools, within the FRF context, is not possible because to date the only ones who have dealt with the subject in literature have limited themselves to using the traditional feature importance measures, e.g. Gini or Shannon index-based mean decrease.

\begin{figure}[htbp]
    \centering
    \includegraphics[width=\textwidth]{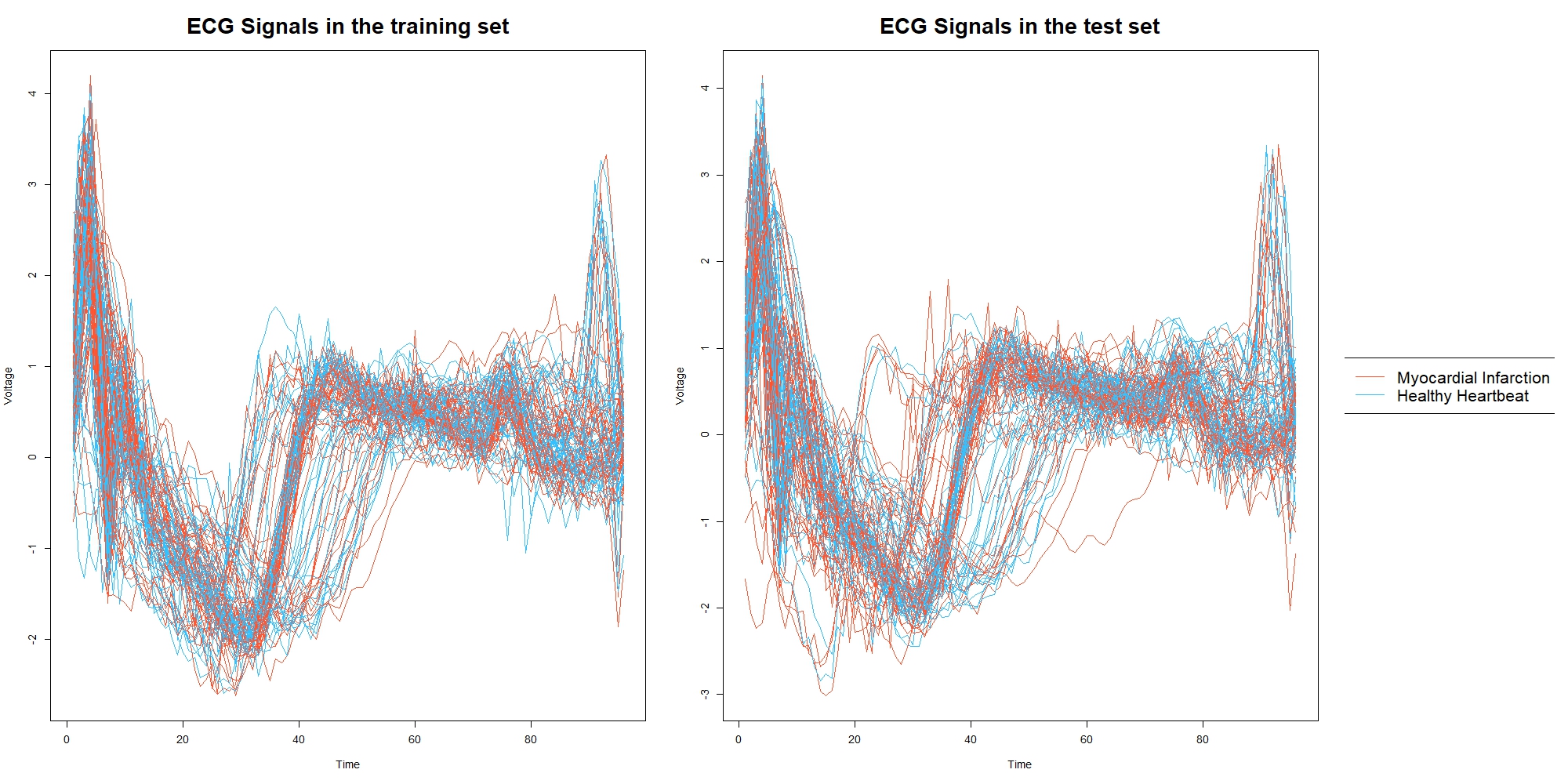}
    \caption{Original ECG signals from the training and test sets. The left panel shows the signals from the training set, and the right panel shows the signals from the test set. The blue curves correspond to healthy patients (NH), while the green curves represent those diagnosed with heart disease (MI).}
    \label{fig:smoothed_signals}
\end{figure}

\begin{figure}[htbp]
    \centering
    \includegraphics[width=\textwidth]{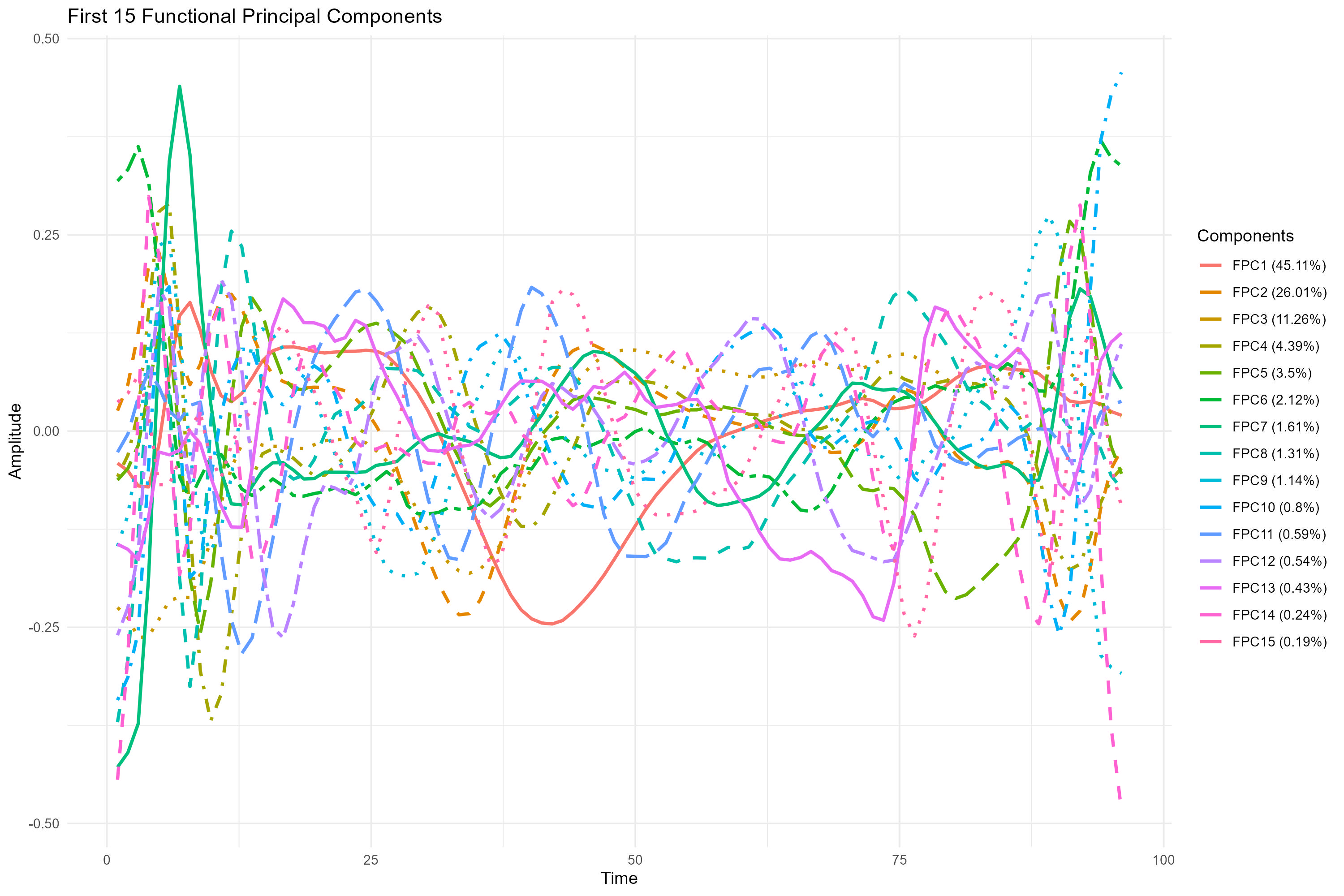}
    \caption{Functional Principal Components Decomposition of the dataset. The plot shows the first 15 Functional Principal Components (FPCs), with the associated explained variance.}
    \label{fig:fpc_decomposition}
\end{figure}

Figure \ref{fig:smoothed_signals} presents the original signals from both the training and test sets. In these figures, the blue curves correspond to signals from healthy patients (NH), while the orange curves represent those diagnosed with heart disease (MI). Figure  \ref{fig:fpc_decomposition} illustrates the first fifteen FPCs. Each curve in the graph represents a different FPC, and the associated legend displays the percentage of total variability explained by each component. The first FPC explains 45.11\% of the variability, while subsequent components explain progressively smaller portions, with the 15th component accounting for 0.19\%.

A common question that arises is: why should we consider all of these FPCs?
In the context of supervised classification it is widely proven that even FPCs that account for very small percentages of the total variability can be crucial.
This is because these minor FPCs may capture subtle but significant variations in the data that are essential for distinguishing between different classes.
Therefore, this analysis does not focus solely on the first few FPCs that collectively explain 70-80\% of the variability, which is often the case in the classical unsupervised dimensionality reduction contexts.
Hence, the analysis deliberately retains a large number of FPCs, focusing not only on the major components but also on those with smaller contributions to the overall variability. Indeed, previous studies (see e.g., \citep{maturo2023supervised}) have shown that often some FPCs (particularly the first) may catch a variability common to all signals, often neglecting sub-patterns, as evidenced also by \citet{maturo2024combining}.

\begin{figure}[htbp]
    \centering
    \includegraphics[width=12cm]{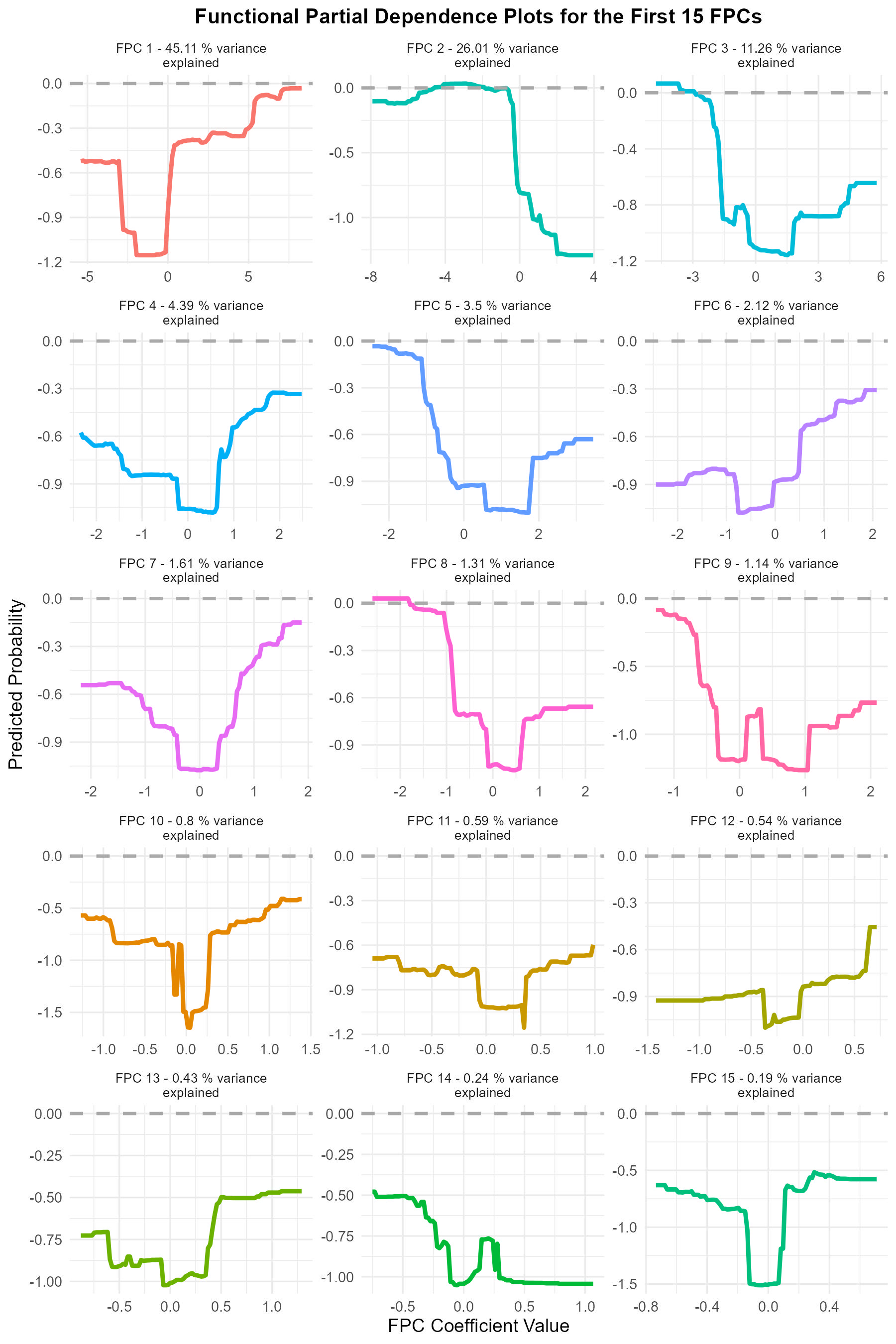}
    \caption{Functional Partial Dependence Plots (FPDPs) for the First 15 Functional Principal Components (FPCs). Each subplot shows the variation in predicted probability as the score of a single FPC is varied, while the other scores remain constant. The plots are ordered by the amount of variance explained by each FPC.}
    \label{fig:pdp_plots}
\end{figure}

Figure  \ref{fig:pdp_plots} shows the first explainability tool of this research: the Functional Partial Dependence Plots (FPDPs).
This study extends the application of PDPs to the FDA context, with a particular emphasis on the time domain, which is crucial in FDA. In FDA, interpreting results within the time domain is essential, making our proposal  necessary for accurately capturing the dynamics of the decision-making process.
In the previous work on FRF and FCTs mentioned above, excellent performance was achieved in terms of accuracy and AUC, but it was not clear whether the total effect of FPCs (or b-spline coefficients) eventually became positive or negative on the outcome variable, how in the time domain, and with what effect in terms of outcome.

FPDPs display the relationship between the predicted probability and the coefficient value of each of the first 15 FPCs. The plots are ordered by the amount of variance explained by each FPC, with the most significant FPCs displayed first. The FPDPs show how varying the score of a single FPC while holding others constant affects the model's predicted probability. This allows us to visualise the importance and effect of each FPC. The FPCs are ordered from the one that explains the most variance to the one that explains the least. The varying shapes of the PDPs suggest that different FPCs contribute differently to the model. Some FPCs show strong nonlinear relationships with the predicted probability, indicating their critical role in capturing essential patterns in the data. This analysis helps understand which FPCs are most influential in the model and how they affect the predictions, providing valuable insights into the model's behavior and the underlying data structure.

In this analysis, we utilise the logit scale for the FPDPs to emphasise changes in the predicted probabilities, mainly when these probabilities are near $0.5$. The logit scale transforms the probability range $[0, 1]$ into the log-odds range, making it more sensitive to variations in predictor variables. This sensitivity is beneficial for detecting subtle effects that may be less apparent on the raw probability scale. By spreading out the middle range of probabilities, the logit scale allows us to gain deeper insights into how specific FPCs influence the classification model's predictions. Consequently, the interpretation of the FPDPs must be made with reference to positivity and negativity, which in our application correspond respectively to a higher probability for the class \textit{desease} and the class \textit{healthy}.

\begin{figure}[htbp]
    \centering
    \includegraphics[width=0.5\textwidth]{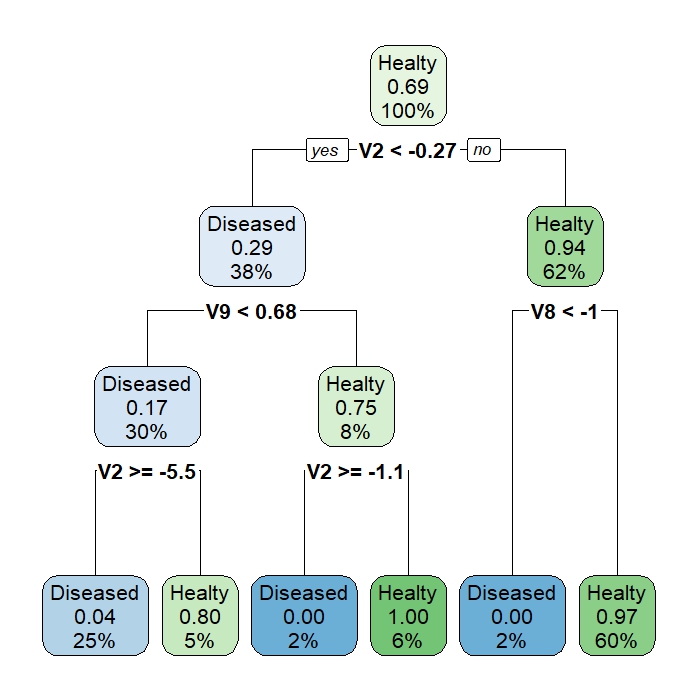}
    \caption{Pruned FCT illustrating the relationship between various features and the predicted health outcome. Each node represents a decision point based on one of the features, leading to a classification of either Healthy or Diseased. The percentage values within each node indicate the proportion of samples classified into that category, while the color intensity reflects the purity of the node's classification.}
    \label{fig:decision_tree}
\end{figure}

Figure \ref{fig:decision_tree} displays a pruned FCT that illustrates the decision-making process of a single base classifier, ultimately leading to the classification of individuals as either \textit{Healthy} or \textit{Diseased}. Each node in the tree represents a split based on the value of a particular FPCs' score, and the branches indicate the direction of the split. The leaves of the FCT show the final classification, with associated probabilities and percentages of the total dataset that fall into each category.
The FCT clearly visualizes how specific feature thresholds determine the classification outcomes. For instance, the root node uses the second FPC to make the initial split, leading to different paths based on whether its score is less than -0.27.

Although in the paper we focus on the FRF and therefore, the interpretation of a single FCT loses meaning because we have an ensemble of FCTs all different and therefore unable to interpret as we have for a single FCT, we show the single FCT for a specific reason. It is exciting to note that a single FCT, trained on the whole dataset (not bootstrap) and all FPCs (not subjected to random extraction at each split), reveals some salient aspects in accordance with the FPDP. The first split of the single FCT takes place on the second FPC for a score value equal to -0.27; the same circumstance is found in the FPDP, proving that the method works well because we note in the FPDP of the second FPC that for values of the scores less or equal to -0.27 there is a high probability of having heart problems. Similarly, as the FCT shows, the same FPC interacts with itself in the single tree; for values greater than -5.5, the impact on the outcome is reduced by returning to the negative zone (tendency towards healthy prediction of the outcome).
Thus, the FCT displayed in Figure \ref{fig:decision_tree} is presented as an illustrative example and to show the correspondence of some intuitions.
However, it is essential to note that the model we use is FRF. If we rely solely on a FCT, the resulting model would be more easily interpretable, providing a clear and straightforward path from input features to classification outcomes. However, a single FCT has significant limitations, such as a higher risk of overfitting and reduced predictive accuracy compared to ensemble methods like FRF. While it offers interpretability, it may fail to capture the full complexity and interactions within the data, leading to less robust and generalisable predictions due also to high variance. 

\begin{figure}[htbp]
    \centering
    \includegraphics[width=\textwidth]{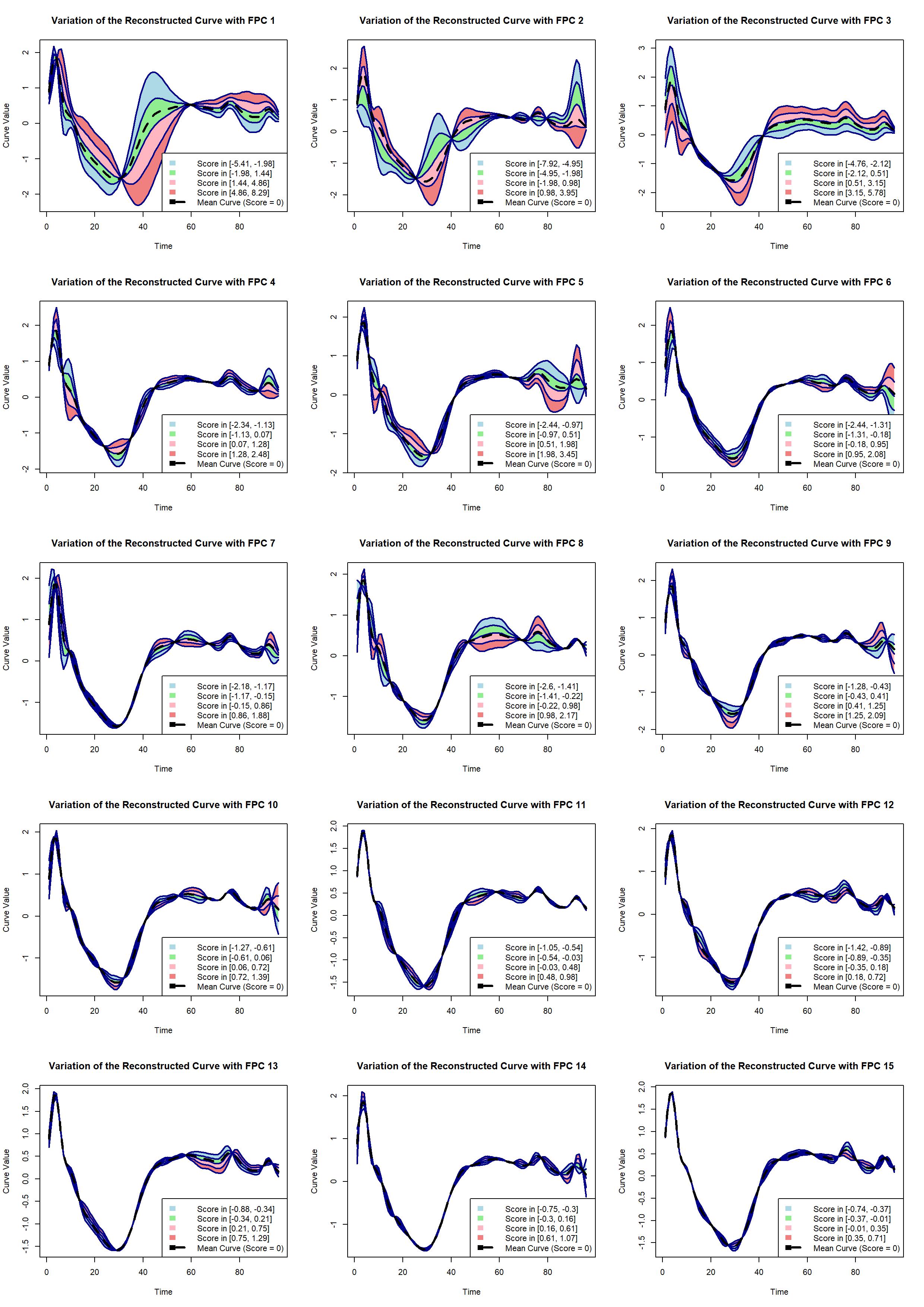}
    \caption{Variation of the Reconstructed Curve with the First Fifteen Functional Principal Components (FPCs). Each subplot shows how variations in the score of a single FPC influence the shape of the reconstructed curve.}
    \label{fig:FPC2dvariation15}
\end{figure}

Nonetheless, while the FPDPs offer valuable insights into the influence of individual FPCs on the model’s predictions, it is crucial to recognise that there can be complex interactions between FPCs that a simple comparison between a single tree and the PDPs cannot fully explain. These interactions may involve multiple FPCs working together in ways that are not immediately apparent from examining individual plots. Thus, while the PDPs are highly useful for enhancing the explainability of the model, they represent only a part of the broader picture. 
Above all, since we are in the FDA domain, the reading of the time domain also assumes great importance, and thus the study of FPDPs must be refined by functional reconstruction through the FPCs in Figure \ref{fig:FPC2dvariation15}.

In Figure \ref{fig:FPC2dvariation15}, the reconstructed curves are visualised to reflect the different intervals of scores for each of the first fifteen FPCs. Each subplot represents the influence of a single FPC (linked to its score) on the shape of the reconstructed curve. The shaded regions highlight the variability (curves' range) within specific scores' intervals, providing a visual stratification of how each FPC affects the functional data reconstruction.
This approach allows for a more straightforward interpretation of the relationship between FPC scores and the resulting curve shapes across different domain regions. The coloured bands in each subplot represent the potential range of curve values corresponding to specific intervals of FPC scores indicated in the legends (the evaluation thresholds are different between the various FPCs' scores because different ranges characterise each, but each is divided into four windows). The more pronounced deviations from the mean curve (indicated by the black dashed line) show the significant impact specific score ranges can have on the curve's amplitude and shape.

Focusing on these shaded areas reveals how each FPC contributes to the overall structure of the functional data. For example, in many subplots, the band representing higher score intervals (in absolute terms) often shows a significant deviation from the mean curve. This indicates that higher scores generally strongly influence the curve's shape, emphasizing or diminishing specific functional features.
Understanding these variations is fundamental for interpreting FPDPs in the time domain and capturing how and how much the scores of specific FPCs influence the predicted probabilities of outcome classes in a supervised classification setting. However, only by closely examining the individual charts in Figure \ref{fig:FPC2dvariation15} can one fully appreciate how the shape of the reconstructed curve drive by different FPC score intervals—affects the model's predictions.

\begin{figure}[htbp]
    \centering
    \begin{subfigure}[b]{0.45\textwidth}
        \centering
        \includegraphics[width=\textwidth]{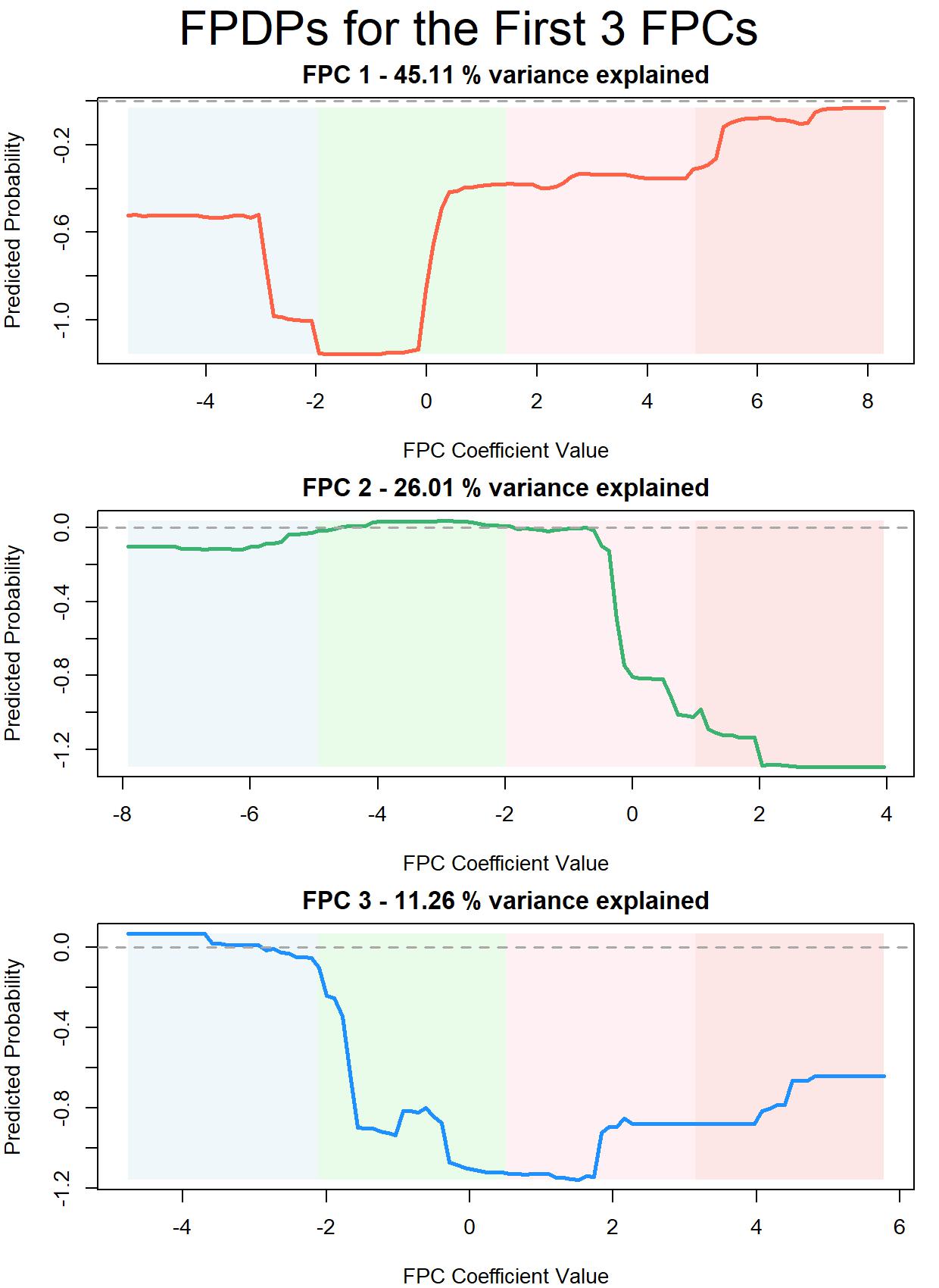}
        \caption{Functional Partial Dependence Plots for the First 3 FPCs}
        \label{fig:pdp3}
    \end{subfigure}
    \hfill
    \begin{subfigure}[b]{0.45\textwidth}
        \centering
        \includegraphics[width=\textwidth]{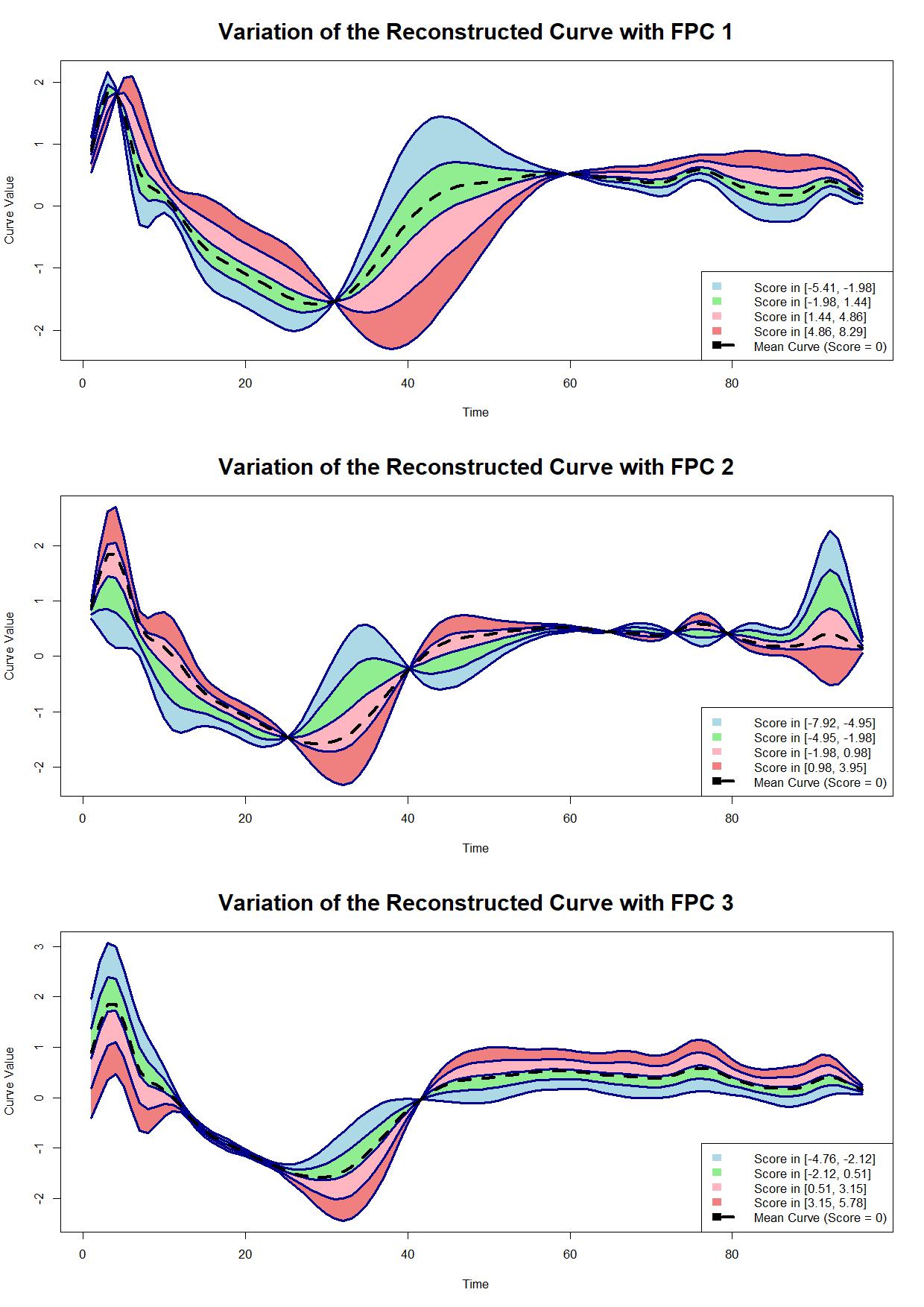}
        \caption{Variation of the Reconstructed Curves with the First 3 FPCs}
        \label{fig:FPCvar}
    \end{subfigure}
    \caption{Comparison of FPDPs and FPC Variation for different scores' intervals.}
    \label{fig:comparison}
\end{figure}

To better understand how to interpret the relationship between FPDPs and functional data reconstructions, we focus closely on the first three FPCs as illustrated in Figure \ref{fig:comparison}. 
This comparison serves as a magnifying lens, helping to clarify the joint reading of PDPs alongside the corresponding reconstructed functional curves. 
In the left column (Figure \ref{fig:pdp3}), the FPDPs show the relationship between each FPC's coefficient value and the predicted outcome probability. These plots help identify how different ranges of FPC scores influence the model's predictions. The colored areas in these PDP plots correspond to the score intervals used in the right column (Figure \ref{fig:FPCvar}), where the reconstructed curves are plotted based on the shapes of FPCs (depending on $t$).
Hence, the right column (Figure \ref{fig:FPCvar}) visualises the effect of varying the FPC scores within specific intervals on the shape of the reconstructed curves in the time domain. The areas in these plots are shaded according to the same intervals shown in the PDPs, providing a clear visual link between the FPC score ranges and their impact on both the range of the reconstructed curve shapes and the model predictions.
This combined approach offers a more comprehensive understanding of how each FPC contributes to the prediction model, particularly in understanding how changes in the FPC scores translate into variations in the functional data, affecting the predicted probabilities. 

Let's focus on the third FPC as an example of interpreting the associated plots in Figure \ref{fig:comparison}. In the FPDP, we observe that within the first interval from the left, there is a significant increase in the probability of a positive outcome, which, in the context of ECGs, indicates a higher likelihood of a heart problems. The crucial question is: What specific shape must an ECG signal have for this to occur?
The answer lies in the corresponding figure on the right, which shows how a generic curve is reconstructed using only the third FPC. This figure reveals that the probability of a positive outcome increases when the score of the third FPC is less than 2.3 (as indicated also by the FPDP). When we reconstruct the curve using scores below 2.3, we observe a marked increase in the ECG values in the early part of the domain.
This suggests that when the third FPC score is very low, the ECG curve exhibits a specific pattern correlating with a higher risk of heart problems (a peak in the first part of the time domain and a lower value in the second half). The same analytical approach and interpretation can be extended to other FPCs and regions of the time domain. This method provides a novel and insightful way to understand the influence of individual FPCs on the overall data structure, offering a more nuanced interpretation of the functional data.

\begin{figure}[htbp]
    \centering
    \includegraphics[width=0.7\textwidth]{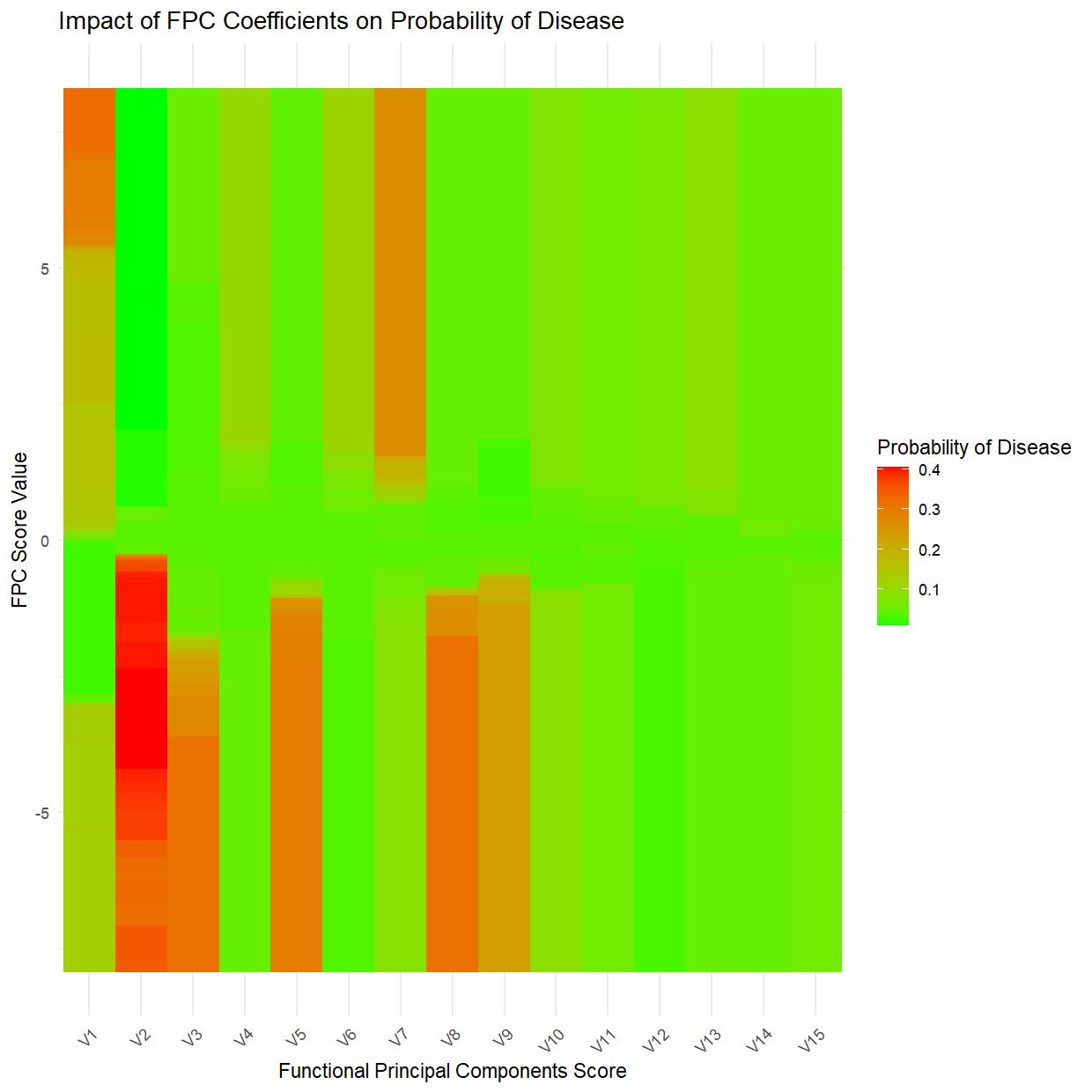}
    \caption{Functional Principal Component Probability Heatmap showing the impact of FPC coefficients on the predicted probability. Green areas indicate a higher probability of being classified as \textit{Healthy}, and red areas indicate a higher probability of being classified as \textit{Diseased}.}
    \label{fig:FPC_Probability_Heatmap}
\end{figure}

Figure \ref{fig:FPC_Probability_Heatmap} demonstrates how FPCs' scores influence the probability of being classified as \textit{Diseased}. The horizontal axis represents the FPCs, while the vertical axis corresponds to the possible scores of these components. The color gradient from green to red indicates the probability of classification: green signifies a higher likelihood of being classified as \textit{Healthy}, whereas red indicates a higher probability of being classified as \textit{Diseased}. This visual tool provides insight into which FPCs and their corresponding score ranges have the most significant impact on the classification outcome, aiding in interpreting the FPC's role in predicting health status.

While this heatmap provides an overview of the impact of each FPC on classification outcomes, for a more nuanced understanding of how these components influence the shape of the functional data over time, it is essential to analyze the plots that reconstruct the original curves using these FPCs. Such analysis, similar to the approach used with FPDPs, allows us to link specific variations in the FPCs to changes in the curve shapes, offering more profound insights into the underlying data patterns and their relationship to the predicted health outcomes.

\begin{figure}[htbp]
    \centering
    \includegraphics[width=1\textwidth]{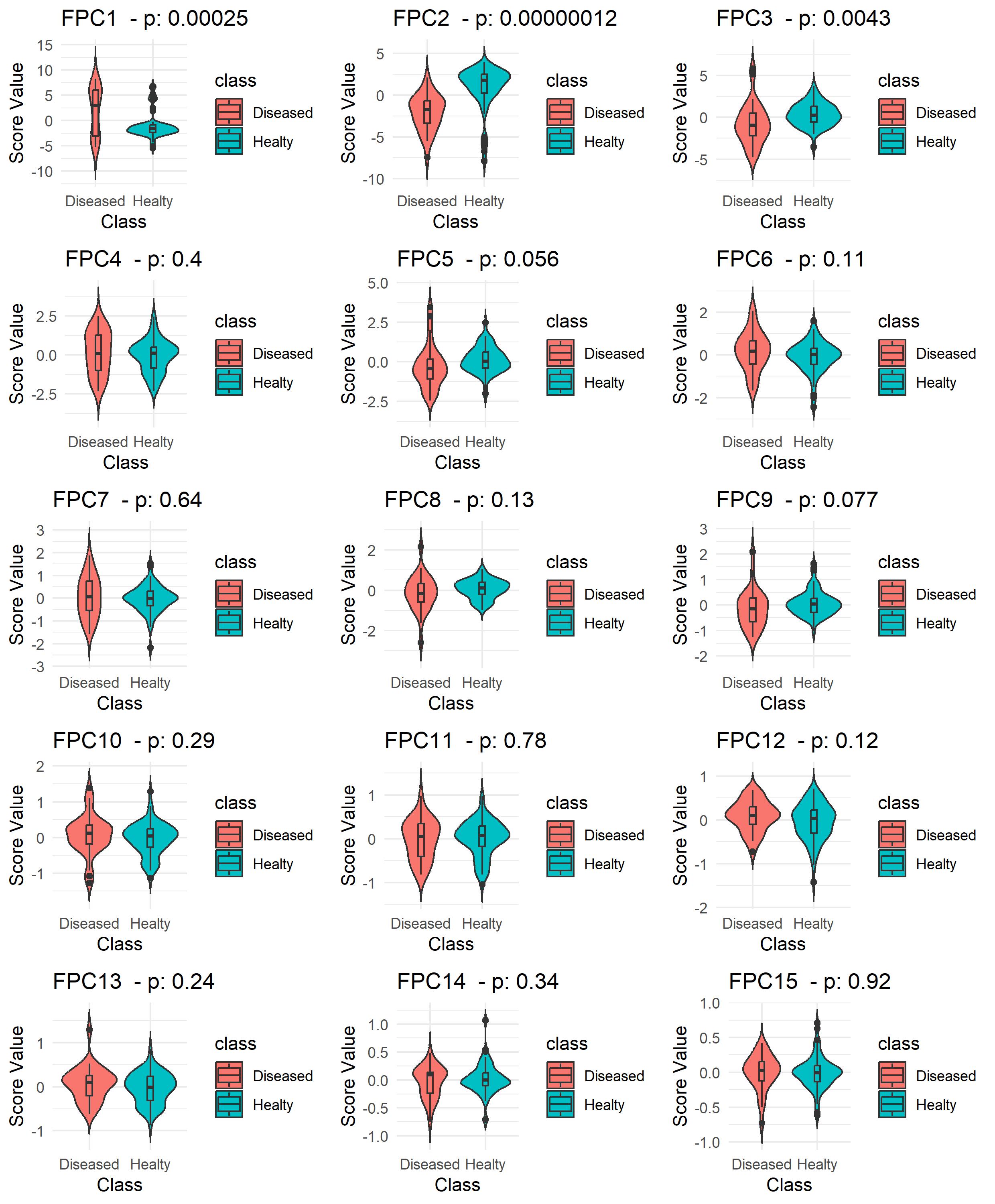}
    \caption{Violin plots of the first 15 FPCs scores by class. The p-values from ANOVA tests are included in the plot titles to indicate the statistical significance of differences between the two outcome classes.}
    \label{fig:fpc_violin_plots}
\end{figure}

Figure \ref{fig:fpc_violin_plots} displays violin plots for the first 15 FPCs scores, divided by class into \textit{Healthy} and \textit{Diseased} groups. The violin plots provide a comprehensive view of the distribution of FPC scores within each class, with box plots included within the violins further to emphasise the median and interquartile range of the data.
The titles of each plot include the p-value from an ANOVA test, indicating the statistical significance of the difference between the score distributions of the two classes. Notably, FPCs with very low p-values, such as FPC1, FPC2, and FPC3, and FPC4 show a significant difference in score distributions between the two classes, suggesting that these components may be critical in distinguishing between \textit{healthy} and \textit{diseased} subjects. Conversely, FPCs with higher p-values indicate less evidence of a significant difference between the two groups.
This plot could already indicate each FPC's positive or negative impact on prediction. However, it is essential to note that this is not guaranteed because the violin plot of the distribution of a single FPC cannot capture complex interactions among multiple FPCs. Nevertheless, also through ANOVA, it serves as an initial visual and inferential indication of noteworthy differences between the groups.

\begin{figure}[htbp]
    \centering
    \includegraphics[width=\textwidth]{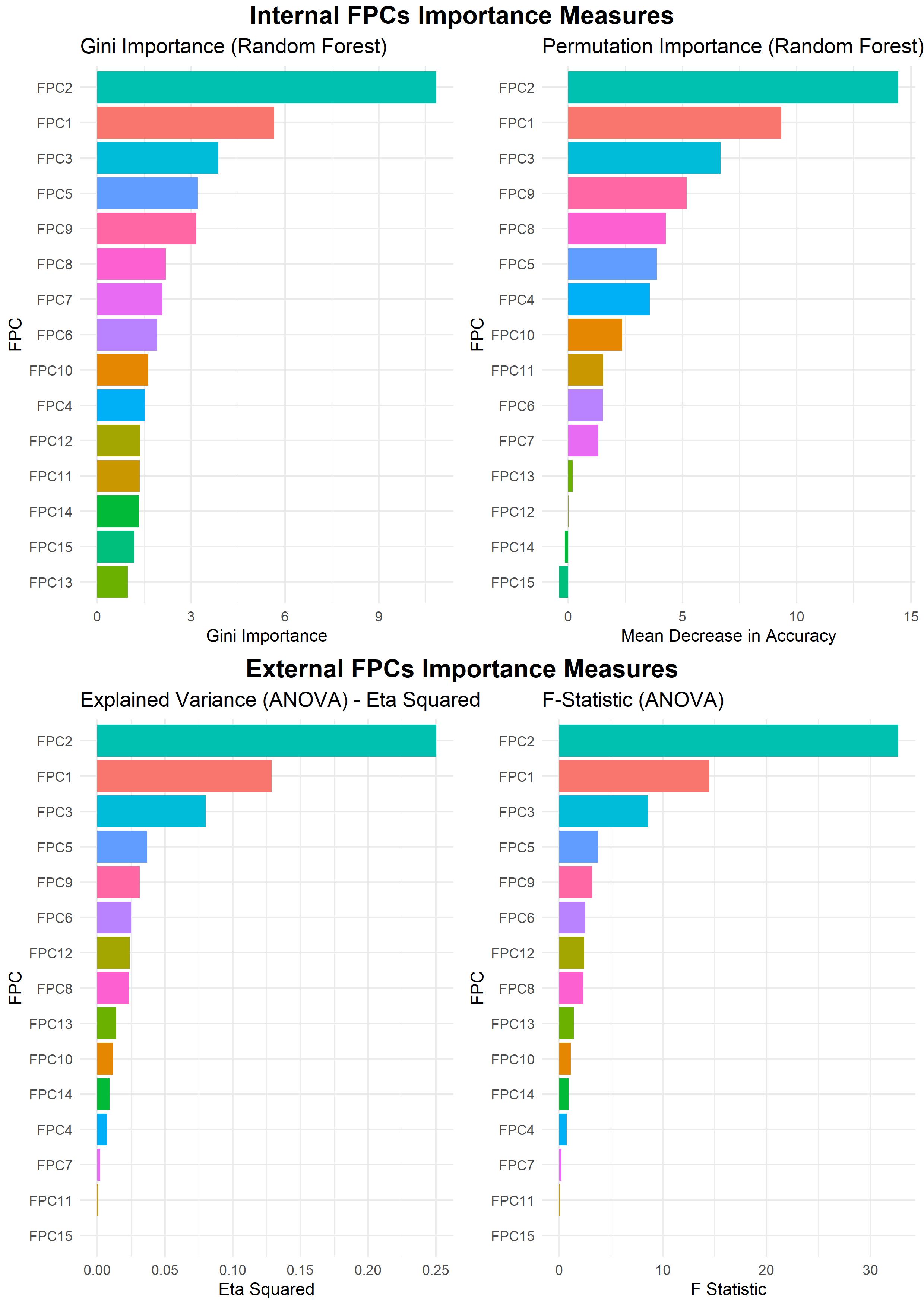}
    \caption{Comprehensive comparison of the importance measures of the first 15 Functional Principal Components (FPCs). The importance measures are divided into internal (model-specific) and external (model-agnostic).}
    \label{fig:FPC_Importance}
\end{figure}

Both internal and external importance measures can assess the role of each FPC in the classification task. The results are visually summarized in Figure~\ref{fig:FPC_Importance}. 
The top row of Figure~\ref{fig:FPC_Importance} displays the internal importance measures derived from the FRF, specifically the Gini Importance and Permutation Importance. These measures reflect the contribution of each FPC to the model's predictive accuracy. Notably, FPC2, FPC1, and FPC3 stand out with the highest importance values, suggesting that these components capture the most significant patterns in the data relevant for predicting the outcome. The Gini Importance and Permutation Importance together provide insight into which FPCs are most influential within the model itself.

On the other hand, it is insightful to complement the internal model-based measures with external measures, which are often overlooked in the literature. External measures, such as Eta Squared and F-Statistics derived from ANOVA, offer an independent evaluation of how much variance in the outcome can be attributed to each FPC, separate from the model’s influence. By comparing these external metrics with internal measures like Gini Importance and Permutation Importance, we can assess whether the model's internal rankings are robust or potentially biased or overly dependent on the specific model architecture. This dual approach helps identify any discrepancies, providing a more balanced and thorough understanding of the FPCs' true importance in the data structure and their predictive power.
Thus, the bottom row of Figure~\ref{fig:FPC_Importance} presents the external importance measures, which include the Eta Squared and F-Statistic obtained from ANOVA tests. Once again, FPC2, FPC1, and FPC3 emerge as the most essential components, underscoring their significant explanatory power beyond the model’s internal mechanics. 
This dual analysis of FPC importance allows us to understand not only which components are crucial for the classification model but also how they relate to the underlying structure of the data. By integrating internal and external measures, we gain a comprehensive view of the functional principal components essential for distinguishing between classes, which is particularly valuable in applications where model explainability is key.
We emphasize that in this case the most important features are the first three but very often this does not happen, especially when the data has such a high variability in the time domain.

\begin{figure}[htbp]
    \centering
    \includegraphics[width=\textwidth]{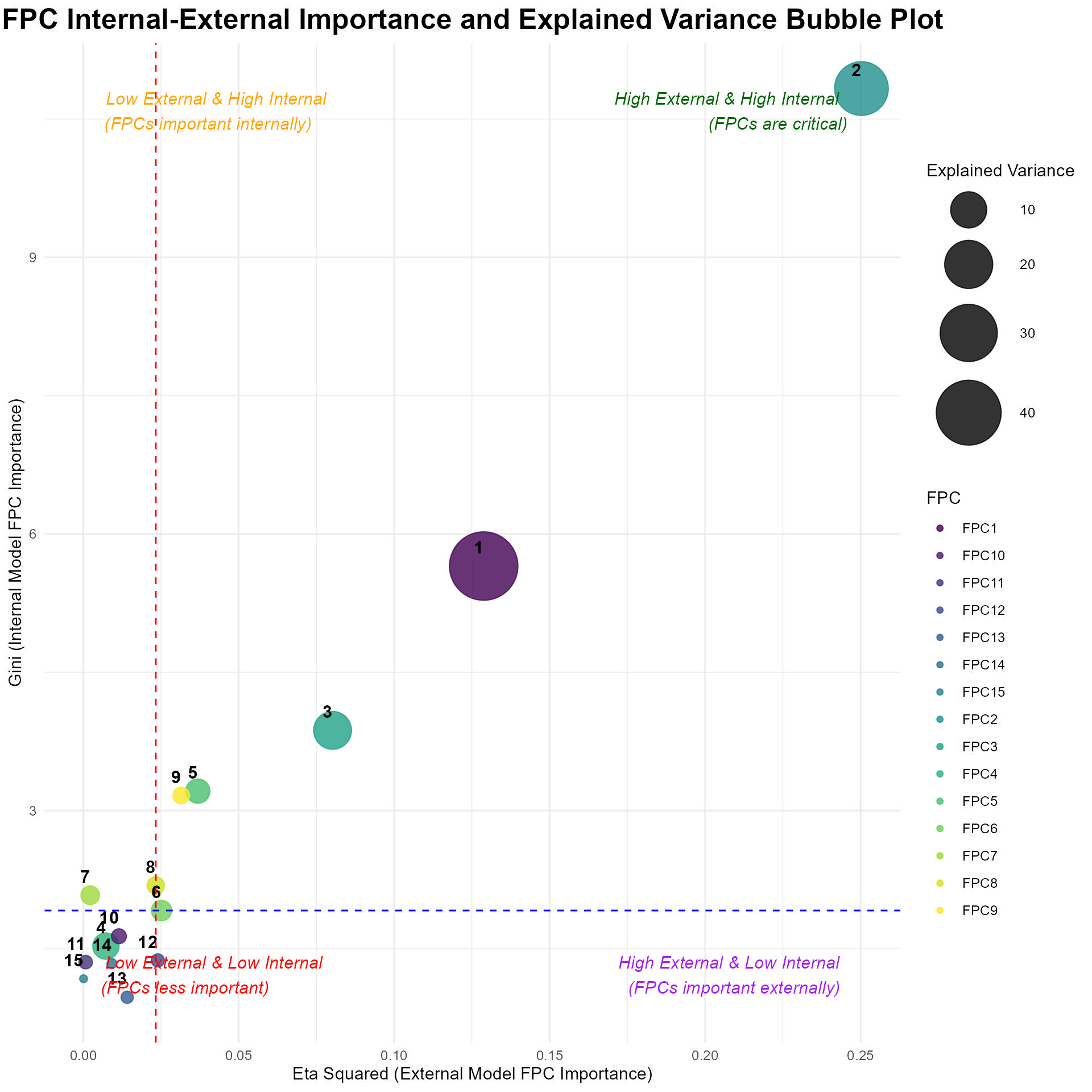}
    \caption{FPC Internal-External Importance and Explained Variance Bubble Plot: Scatter plot of FPCs based on internal and external importance measures. The X-axis shows the external importance (Eta Squared), and the Y-axis represents the internal importance (Gini). The size of the bubbles corresponds to the explained variance of each FPC.}
    \label{fig:fpc_importance.plot}
\end{figure}

Finally, Figure  \ref{fig:fpc_importance.plot} illustrates the importance of FPCs based on internal and external importance measures, i.e. the FPCs Internal-External Importance and Explained Variance Bubble Plot.
For this plot, we chose eta-squared as a model-agnostic measure and the Gini index as a model-specific measure, with the explained variance represented by the size of the bubbles.
The X-axis represents the Eta Squared value, which reflects the external model importance derived from ANOVA. The Y-axis shows the Gini index, a measure of internal model importance derived from the FRF. The size of each bubble corresponds to the explained variance of each FPC, while the color denotes the FPC identity.
The plot is divided into four quadrants by dashed lines representing the median internal and external importance values. The top right quadrant indicates FPCs that are highly important both internally and externally, making them critical components. The top left quadrant shows FPCs that are mainly important internally, while the bottom right quadrant identifies FPCs with external importance but less internal significance. Finally, the bottom left quadrant represents FPCs that are less important in both aspects.

Understanding the gaps between model-specific and model-agnostic FPCs importance is 
essential because a feature that appears highly important within the model (high internal importance) might have a negligible effect when evaluated independently (low external importance). This could indicate that the model may overvalue the feature due to interactions with other variables or overfitting. 
This happens, for example, with the FPC7, which shows a value above the median in terms of model-specific importance, but the model-agnostic measures slightly underestimate this importance.
FPC2 stands out as a critical component with high importance both internally and externally. It explains a significant amount of variance, suggesting that FPC2 is a robust component, essential both within the model context and in broader analysis.
FPC1 shows high internal but significantly lower external importance despite explaining much of the total variance. This might indicate that FPC1 is strongly influenced by the specific model used and may not be as relevant in a more general context.
While FPC3 is less important internally than FPC1, it has decent external importance, suggesting a balance between its contribution to the model and general relevance.
Although less critical, other FPCs, like FPC9, still contribute to the model's predictive power in different ways.

\section{Discussion and Conclusions}

In the era of big data, investigating high-dimensional datasets has become a crucial challenge across various domains, including medicine, ecology, and economics. The exponential growth in data availability offers exceptional opportunities for gaining insights but also raises significant analytical challenges, particularly when grappling with the curse of dimensionality. As the number of dimensions increases, the volume of data points grows exponentially, complicating extracting meaningful patterns and relationships using traditional statistical approaches. FDA has emerged as a powerful solution to these challenges, transforming high-dimensional data into functional forms capable of catching complex temporal and spatial patterns in a more coherent and interpretable manner \citep{Ferraty2003, Ramsay1991}. This capability has made FDA a robust tool in various applications, particularly in biomedical fields, where continuous monitoring data like electrocardiogram (ECG) signals are prevalent \citep{Maturo2022SIM}.

One of the most promising methods for the supervised classification of functional data is the FRF. This approach combines the flexibility of random forests with the strengths of FDA, allowing for effective handling of high-dimensional functional data \citep{Maturo2022SIM, maturo2023supervised}. However, despite the demonstrated effectiveness of FRFs in predictive tasks, a significant drawback persists: the need for more transparency and explainability. Like many machine learning models, FRFs often function as black boxes, making it challenging to comprehend individual features contribution.

The implication of transparency in machine learning models is increasingly recognized, especially in critical fields such as healthcare and environmental science, where model predictions can have significant real-world consequences \citep{Maturo2022SIM}. To address this need, our paper introduces a novel set of explainability tools designed to enhance the interpretability of FRF. These tools aim to bridge the gap between model accuracy and understanding, providing users with the means to demystify the decision-making process of FRFs.

The key contributions of this work include the development of \textit{Functional Partial Dependence Plots (FPDPs)}, which illustrate the influence of individual FPCs on the model's predictions, and \textit{FPC Probability Heatmaps}, which visualize how changes in FPC scores affect the predicted probabilities for different classes. Additionally, we introduce various FPC importance metrics, including the \textit{FPC Internal-External Importance and Explained Variance Bubble Plot}, which offers a comprehensive view of the significance of each FPC from both internal and external perspectives.

These methods are demonstrated through an application to an ECG dataset, a classic example of functional data where the shape and variability of heart activity curves are crucial for accurate diagnosis. Applying these explainability tools reveals deep insights into how individual FPCs influence the classification of ECG signals, thereby enhancing the interpretability of the FRF model. 
To the best of our knowledge, this study is the first to specifically address the issue of explainability within the context of FRF. Prior works, including those by Maturo and Verde, have primarily focused on the interpretability of individual Functional Classification Trees (FCTs), without delving into the explainability of ensemble methods like FRFs \citep{Maturo2022SIM, Maturo2022CS_FRF}. Existing literature predominantly discusses traditional variable importance measures, but more is needed to explain the intricacies of FRFs. As a result, there needs to be more alternative methods in this domain, justifying the absence of references to other approaches.
Moreover, this paper intentionally avoids discussing accuracy or other performance metrics, as these aspects are not the focus of this study. The predictive power of ensemble methods for functional data has already been well-documented in the literature, particularly in the works of  \citet{Maturo2022SIM, Maturo2022CS_FRF}. Thus, this study does not engage in performance comparisons, as it lies beyond the scope of our objectives.

In conclusion, this paper significantly contributes to Functional Data Analysis by addressing a critical gap in the explainability of Functional Random Forests. The novel explainability tools introduced here provide a pathway to greater transparency, making these powerful models more accessible and trustworthy for practitioners across various domains. Future research could explore further enhancements to these tools or their application to other complex datasets, thereby improving the transparency and usability of machine learning models in high-dimensional functional data contexts.

\section*{Declarations}

\subsection*{Funding and/or Conflicts of interests/Competing interests}

All the authors declare that they did not receive support from any organisation for the submitted work.
All authors certify that they have no affiliations with or involvement in any organization or entity with any financial or non-financial interest in the subject matter or materials discussed in this manuscript.

\subsection*{Use of generative AI in scientific writing}

While preparing this work, the authors used \textit{Grammarly AI} to improve the English language. While using this tool, the authors reviewed and edited the content as needed and took full responsibility for the publication's content.

\bibliography{_biblio}

\end{document}